\documentclass[runningheads]{llncs}
\usepackage{graphicx}
\usepackage{comment}
\usepackage{amsmath,amssymb} %
\usepackage{color}
\usepackage{color}
\usepackage{amsmath}
\usepackage{wrapfig}
\usepackage{array}
\usepackage{newlfont} %
\usepackage{textcomp}
\usepackage{multirow}
\usepackage{hhline}
\usepackage{tabularx}
\usepackage{bigstrut}
\usepackage{afterpage}
\usepackage{algorithmic}
\usepackage{microtype}
\usepackage{mathtools}
\usepackage{authblk}
\usepackage{booktabs} %
\usepackage[ruled]{algorithm2e} %
\usepackage{cite}

\newcolumntype{?}{!{\vrule width 1pt}}

\newcommand{\Caption}[2]{\caption[#1]{{\footnotesize #1} {\footnotesize #2}}}

\definecolor{DeltaColor}{rgb}{0.039,0.73,0.71}
\definecolor{SigmaColor}{rgb}{0.98,0.45,0.0}
\definecolor{WeikaiColor}{rgb}{0.8,0,0}
\definecolor{AlphaColor}{rgb}{0,0,0.8}
\definecolor{BetaColor}{rgb}{0.8,0,0.8}
\definecolor{GammaColor}{rgb}{0.514,0.34,0.224}
\definecolor{EpsilonColor}{rgb}{0.353,0.725,0.906}
\definecolor{PurpleColor}{rgb}{0.5,0,0.7}
\definecolor{OrangeColor}{rgb}{1,0.5,0}

\newcommand{\nothing}[1]{}

\definecolor{AudioColor}{rgb}{0.56,0.34,0.62}

\definecolor{DeadlineColor}{rgb}{0.9,0.4,0} %

\definecolor{figred}{rgb}{1,0,0}
\definecolor{figgreen}{rgb}{0,0.6,0}
\definecolor{figblue}{rgb}{0,0,1}
\definecolor{figpink}{rgb}{1,0.63,0.63}

\newcounter{pccount}
\newcommand{\filename}[1]{\url{#1}}
\newcommand{\foldername}[1]{\url{#1}}

\makeatletter
\def\thickhline{%
  \noalign{\ifnum0=`}\fi\hrule \@height \thickarrayrulewidth \futurelet
   \reserved@a\@xthickhline}
\def\@xthickhline{\ifx\reserved@a\thickhline
               \vskip\doublerulesep
               \vskip-\thickarrayrulewidth
             \fi
      \ifnum0=`{\fi}}
\makeatother

\newlength{\thickarrayrulewidth}
\makeatletter
\newcommand{\printfnsymbol}[1]{%
  \textsuperscript{\@fnsymbol{#1}}%
}
\makeatother

\newcommand{\usc}{University of Southern California}
\newcommand{\uscict}{USC Institute for Creative Technologies}

\newcommand{\pinscreen}{Pinscreen}

\begin{document}
\pagestyle{headings}
\mainmatter
\def\ECCVSubNumber{4214}  %

\title{Monocular Real-Time Volumetric Performance Capture} %

\titlerunning{Monocular Real-Time Volumetric Performance Capture}
\author{
Ruilong Li \inst{1,2} \thanks{indicates equal contribution}
\and Yuliang Xiu \inst{1,2} \printfnsymbol{1} 
\and Shunsuke Saito \inst{1,2}
\and Zeng Huang \inst{1,2}\\
\and Kyle Olszewski \inst{1,2}
\and Hao Li \inst{1,2,3}
}
\institute{\usc \and \uscict \and \pinscreen\\
\email{\{ruilongl, yxiu, zenghuan\}@usc.edu, \newline \{shunsuke.saito16,
olszewski.kyle\}@gmail.com, hao@hao-li.com}}
\authorrunning{R. Li et al.}

\maketitle

\begin{abstract}
We present the first approach to volumetric performance capture and novel-view rendering at real-time speed from monocular video, eliminating the need for expensive multi-view systems or cumbersome pre-acquisition of a personalized template model. Our system reconstructs a fully textured 3D human from each frame by leveraging Pixel-Aligned Implicit Function (PIFu). While PIFu achieves high-resolution reconstruction in a memory-efficient manner, its computationally expensive inference prevents us from deploying such a system for real-time applications. To this end, we propose a novel hierarchical surface localization algorithm and a direct rendering method without explicitly extracting surface meshes. By culling unnecessary regions for evaluation in a coarse-to-fine manner, we successfully accelerate the reconstruction by two orders of magnitude from the baseline without compromising the quality. Furthermore, we introduce an Online Hard Example Mining (OHEM) technique that effectively suppresses failure modes due to the rare occurrence of challenging examples. We adaptively update the sampling probability of the training data based on the current reconstruction accuracy, which effectively alleviates reconstruction artifacts. Our experiments and evaluations demonstrate the robustness of our system to various challenging angles, illuminations, poses, and clothing styles. We also show that our approach compares favorably with the state-of-the-art monocular performance capture. Our proposed approach removes the need for multi-view studio settings and enables a consumer-accessible solution for volumetric capture.

\end{abstract}

\setlength{\textfloatsep}{0.1cm}

\section{Introduction}

Videoconferencing using a single camera is still the most common approach face-to-face communication over long distances, despite recent advances in virtual and augmented reality and 3D displays that allow for far more immersive and compelling interaction.
The reason for this is simple: convenience.
Though the technology exists to obtain high-fidelity digital representations of one's specific appearance that can be rendered from arbitrary viewpoints, existing methods to capture and stream this data \cite{collet2015high,dou2016fusion4d,orts2016holoportation,guo2019relightables,tang2018real} require cumbersome capture technology, such as a large number of calibrated cameras or depth sensors, and the expert knowledge to install and deploy these systems.
Videoconferencing, on the other hand, simply requires a single video camera, such as those found on common consumer devices, \textit{e.g.} laptops and smartphones.
Thus, if we can capture a complete model of a person's unique appearance and motion from a single consumer-grade camera, we can bridge the gap preventing novice users from engaging in immersive communication in virtual environments.

However, successful reconstruction of not only the geometry but also the texture of a person from a single viewpoint poses significant challenges due to depth ambiguity, changing topology, and severe occlusions. To address these challenges, data-driven approaches using high-capacity deep neural networks have been employed, demonstrating significant advances in the fidelity and robustness of human modeling \cite{varol18_bodynet,natsume2019siclope,Zheng_2019_ICCV,saito2019pifu}. In particular, Pixel-Aligned Implicit Function (PIFu) \cite{saito2019pifu} achieves fully-textured reconstructions of clothed humans with a very high resolution that is infeasible with voxel-based approaches. On the other hand, the main limitation of PIFu is that the subsequent reconstruction process is not fast enough for real-time applications: given an input image, PIFu densely evaluates 3D occupancy fields, from which the underlining surface geometry is extracted using the Marching Cubes algorithm \cite{lorensen1987marching}. After the surface mesh reconstruction, the texture on the surface is inferred in a similar manner. Finally, the colored meshes are rendered from arbitrary viewpoints. The whole process takes tens of seconds per object when using a $256^3$ resolution.
Our goal is to achieve such fidelity and robustness with the highly efficient reconstruction and rendering speed for real-time applications.

To this end, we introduce a novel surface reconstruction algorithm, as well as a direct rendering method that does not require extracting surface meshes for rendering. The newly introduced surface localization algorithm progressively queries 3D locations in a coarse-to-fine manner to construct 3D occupancy fields with a smaller number of points to be evaluated. We empirically demonstrate that our algorithm retains the accuracy of the original reconstruction, while being two orders of magnitude faster than the brute-force baseline. Additionally, combined with the proposed surface reconstruction algorithm, our implicit texture representation enables direct novel-view synthesis without geometry tessellation or texture mapping, which halves the time required for rendering. As a result, we enable $15$ fps processing time with a $256^3$ spatial resolution for volumetric performance capture.

In addition, we present a key enhancement to the training method of \cite{saito2019pifu} to further improve the quality and efficiency of reconstruction. To suppress failure cases that rarely occur during training due to the unbalanced data distribution with respect to viewing angles, poses and clothing styles, we introduce an adaptive data sampling algorithm inspired by the Online Hard Example Mining (OHEM) method \cite{shrivastava2016training}. We incrementally update the sampling probability based on the current prediction accuracy to train more frequently with hard examples without manually selecting these samples. We find this automatic sampling approach highly effective for reducing artifacts, resulting in state-of-the-art accuracy.%

Our main contributions are:
\begin{itemize}
    \item The first approach to full-body performance capture at real-time speed from monocular video not requiring a template. From a single image, our approach reconstructs a fully textured clothed human under a wide range of poses and clothing types without topology constraints.
    \item A progressive surface localization algorithm that makes surface reconstruction two orders of magnitude faster than the baseline without compromising the reconstruction accuracy, thus achieving a better trade-off between speed and accuracy than octree-based alternatives.
    \item A direct rendering technique for novel-view synthesis without explicitly extracting surface meshes, which further accelerates the overall performance. 
    \item An effective training technique that addresses the fundamental imbalance in synthetically generated training data. Our Online Hard Example Mining method significantly reduces reconstruction artifacts and improves the generalization capabilities of our approach.
\end{itemize}

\section{Related Work}
\label{sec:related}

\paragraph{Volumetric Performance Capture}
Volumetric performance capture has been widely used to obtain human performances for free-viewpoint video \cite{kanade1997virtualized} or high-fidelity geometry reconstruction \cite{Vlasic:2009:DSC}. To obtain the underlining geometry with an arbitrary topology, performance capture systems typically use general cues such as silhouettes \cite{matusik2000image,starck2007surface,waschbusch2005scalable,collet2015high}, mutli-view correspondences \cite{kanade1997virtualized,furukawa2010accurate}, and reflectance information \cite{Vlasic:2009:DSC}. While these approaches successfully reconstruct geometry with an arbitrary topology, they require a large number of cameras with accurate calibration and controlled illumination. Another approach is to leverage commodity depth sensors to directly acquire 3D geometry. Volumetric fusion approaches have been used to jointly optimize for the relative 3D location and 3D geometry, incrementally updated from the captured sequence using a single depth sensor in real-time \cite{izadi2011kinectfusion,newcombe2011kinectfusion}. Later, this incremental geometry update was extended to non-rigidly deforming objects \cite{NewcombeFS15,innmann2016volumedeform} and joint optimization with reflectance \cite{guo2017real}. While these approaches do not require a template or category-specific prior, they only support relatively slow motions. Multi-view systems combined with depth sensors significantly improve the fidelity of the reconstructions \cite{orts2016holoportation,dou2016fusion4d,collet2015high} and both hardware and software improvements further facilitate the trend of high-fidelity volumetric performance capture \cite{kowdle2018need,guo2019relightables}. However, the hardware requirements make it challenging to deploy these systems for non-professional users.

\paragraph{Template-based Performance Capture}
To relax the constraints of traditional volumetric performance capture, one common approach is to use a template model as an additional prior. Early works use a precomputed template model to reduce the number of viewpoints \cite{de2008performance, wu2013set} and improve the reconstruction quality \cite{vlasic2008articulated}. Template models are also used to enable performance capture from RGBD input \cite{ye2012performance,zhang2014leveraging}. However, these systems still rely on well-conditioned input from multiple viewpoints. Instead of a personalized template model, articulated morphable models such as SCAPE \cite{anguelov2005scape} or SMPL \cite{loper2015smpl} are also widely used to recover human pose and shapes from video input \cite{guan2009estimating}, a single image \cite{bogo2016keep,lassner2017unite}, or RGBD input \cite{yu2018doublefusion,zheng2018hybridfusion}. More recently, components corresponding to hands \cite{MANO:SIGGRAPHASIA:2017} and faces \cite{cao2013facewarehouse,li2017learning} were incorporated into a body model to perform more holistic performance capture from multi-view input \cite{joo2018total}, which was later extended to monocular input as well \cite{xiang2019monocular,pavlakos2019expressive}. Although the use of a parameteric model greatly eases the ill-posed nature of monocular performance capture, the lack of personalized details such as clothing and hairstyles severely impairs the authenticity of the captured performance. Recently Xu et al. \cite{Xu:2018:MHP} demonstrated that articulated personalized avatars can be tracked from monocular RGB videos by incorporating inferred sparse 2D and 3D keypoints \cite{mehta2017vnect}. The most relevant work to our approach is \cite{habermann2019livecap}, which is the real-time extension of \cite{Xu:2018:MHP} with the reconstruction fidelity also improved with an adaptive non-rigidity update. Unlike the aforementioned template-based approaches, our method is capable of representing personalized details present in the input image without any preprocessing, as our approach is based on a template-less volumetric representation, enabling topological updates and instantaneously changing the subject.

\begin{wrapfigure}[20]{r}{0.35\textwidth}
\centering
\includegraphics[width=0.35\textwidth]{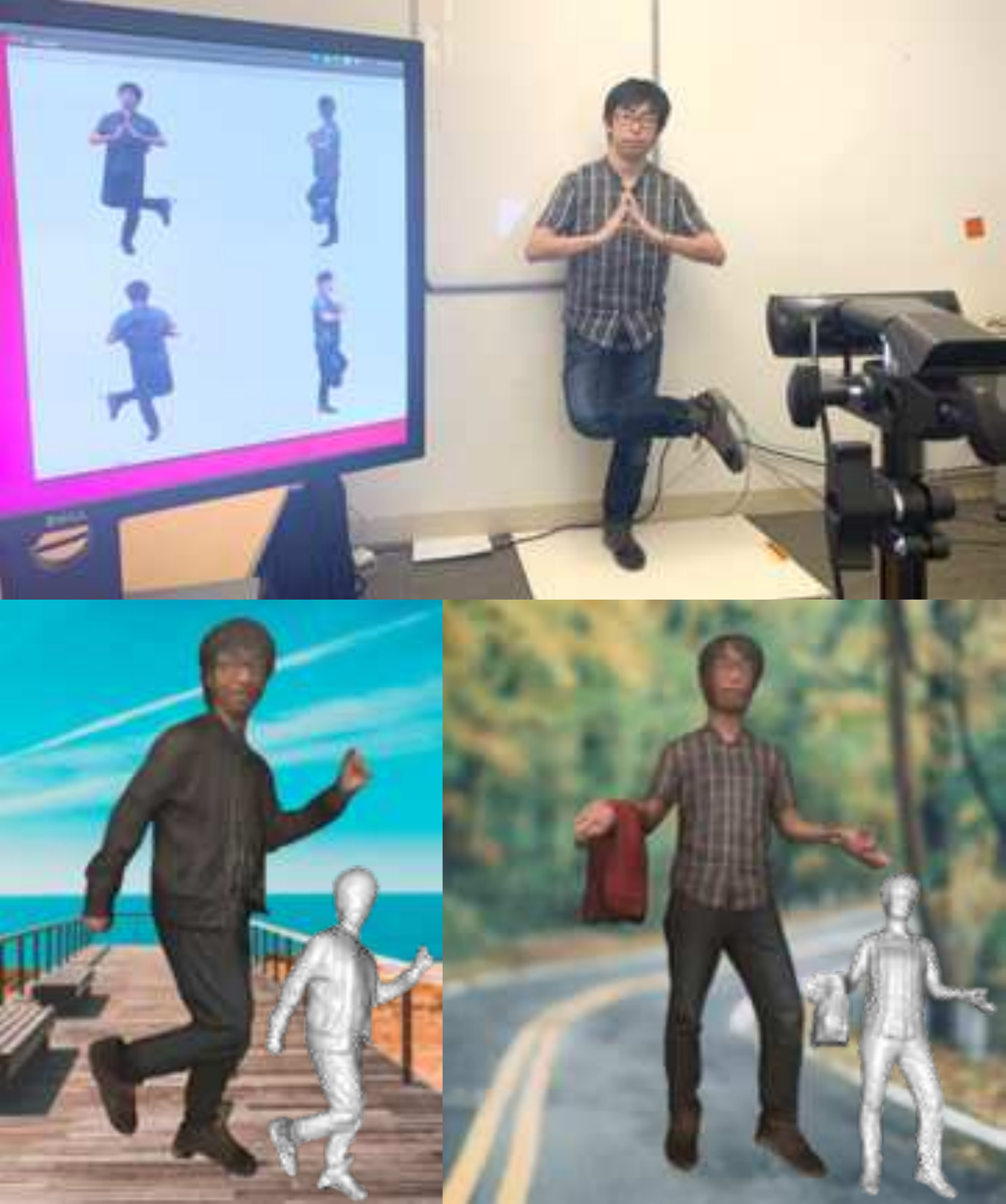}
\caption{A performance captured and re-rendered system in real-time from a monocular input video.}
\label{fig:teaser}
\end{wrapfigure}

\paragraph{Deep Learning for Human Modeling}
To infer fine-grained 3D shape and appearance from unconstrained images, where designing hand-crafted features is non-trivial, we need a high-capacity machine learning algorithm. The advent of deep learning showed promise by eliminating the need for hand-crafted features and demonstrated groundbreaking performance for human modeling tasks in the wild \cite{mehta2017vnect,kanazawa2018end,alp2018densepose}. Fully convolutional neural networks have been used to infer 3D skeletal joints from a single image \cite{popa2017deep,mehta2017vnect,Rogez:2018:LCR-Net}, which are used as building blocks for monocular performance capture systems \cite{Xu:2018:MHP,habermann2019livecap}. 
For full-body reconstruction from a single image, various data representations have been explored, including meshes \cite{kanazawa2018end,kolotouros2019spin}, dense correspondences \cite{alp2018densepose}, voxels \cite{varol18_bodynet,jackson2018human,Zheng_2019_ICCV}, silhouettes \cite{natsume2019siclope}, and implicit surfaces \cite{saito2019pifu,saito2020pifuhd, huang2020arch}. 
Notably, deep learning approaches using implicit shape representations have demonstrated significantly more detailed reconstructions by eliminating the discretization of space \cite{chen2018implicit_decoder,park2019deepsdf,mescheder2018occupancy}. Saito et al. \cite{saito2019pifu} further improve the fidelity of reconstruction by combining fully convolutional image features with implicit functions, and demonstrate that these implicit field representations can be extended to continuous texture fields for effective 3D texture inpainting without relying on precomputed 2D parameterizations. However, the major drawback of these implicit representations is that the inference is time-consuming due to the dense evaluation of the network in 3D space, which prevents its use for real-time applications. Though we base our 3D representation on \cite{saito2019pifu} for high-fidelity and memory-efficient 3D reconstruction, our novel surface inference and rendering algorithms significantly accelerate the reconstruction and visualization of the implicit surface.

\begin{figure}[t]
 \centering{
 \includegraphics[width=.9\linewidth]{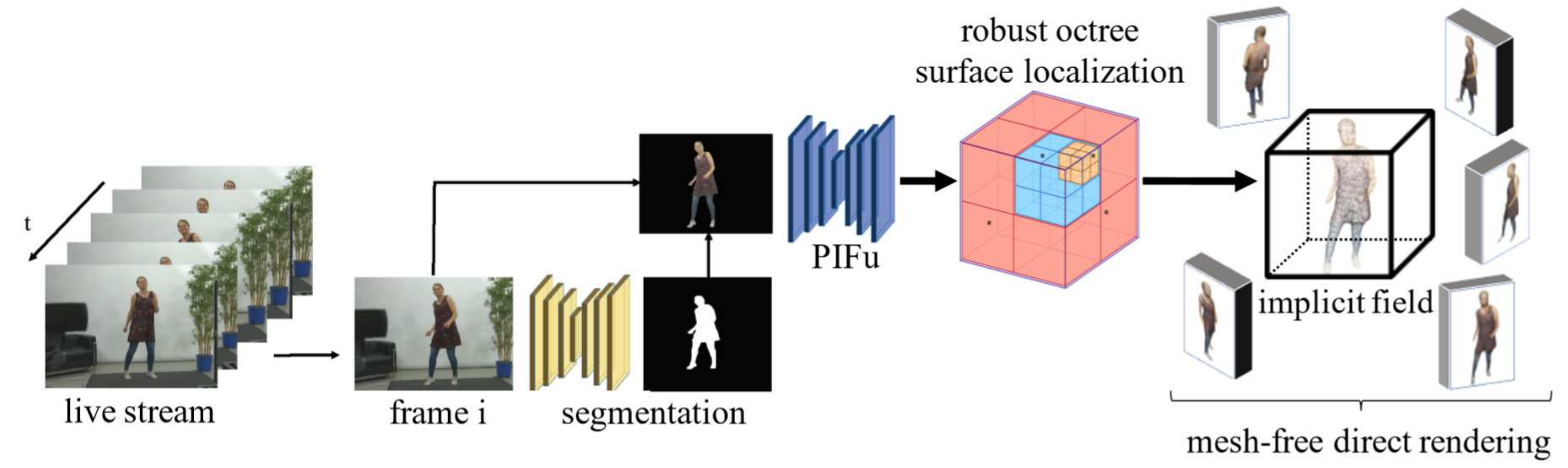}
 \Caption{System overview.}{}
 \label{fig:overview_testing}}
\end{figure}

\section{Method}
\label{sec:method}

In this section, we describe the overall pipeline of our algorithm for real-time volumetric capture (Fig. \ref{fig:overview_testing}). Given a live stream of RGB images, our goal is to obtain the complete 3D geometry of the performing subject in real-time with the full textured surface, including unseen regions. To achieve an accessible solution with minimal requirements, we process each frame independently, as tracking-based solutions are prone to accumulating errors and sensitive to initialization, causing drift and instability \cite{newcombe2011kinectfusion,zollhofer2014real}. Although recent approaches have demonstrated that the use of anchor frames \cite{beeler2011high,dou2016fusion4d} can alleviate drift, ad-hoc engineering is still required to handle common but extremely challenging scenarios such as changing the subject. 

For each frame, we first apply real-time segmentation of the subject from the background. The segmented image is then fed into our enhanced Pixel-Aligned Implicit Function (PIFu) \cite{saito2019pifu} to predict continuous occupancy fields where the underlining surface is defined as a $0.5$-level set. Once the surface is determined, texture inference on the surface geometry is also performed using PIFu, allowing for rendering from any viewpoint for various applications. As this deep learning framework with effective 3D shape representation is the core building block of the proposed system, we review it in Sec. \ref{sec:pifu}, describe our enhancements to it, and point out the limitations on its surface inference and rendering speed. At the heart of our system, we develop a novel acceleration framework that enables real-time inference and rendering from novel viewpoints using PIFu (Sec. \ref{sec:inference}). Furthermore, we further improve the robustness of the system by sampling hard examples on the fly to efficiently suppress failure modes in a manner inspired by Online Hard Example Mining \cite{shrivastava2016training} (Sec. \ref{sec:train}).

\subsection{Pixel-Aligned Implicit Function (PIFu)}
\label{sec:pifu}

In volumetric capture, 3D geometry is represented as the level set surface of continuous scalar fields. That is, given an input frame $\mathbf{I}$, we need to determine whether a point in 3D space is inside or outside the human body. While this can be directly regressed using voxels, where the target space is explicitly discretized \cite{varol18_bodynet,jackson2018human}, the Pixel-Aligned Implicit Function (PIFu) models a function $O(\mathbf{P})$ that queries any 3D point and predicts the binary occupancy field in normalized device coordinates $\mathbf{P} = (P_x, P_y, P_z) \in \mathbb{R}^3$. Notably, with this approach no discretization is needed to infer 3D shapes, allowing reconstruction at arbitrary resolutions.

PIFu first extracts an image feature obtained from a fully convolutional image encoder $g_O(\mathbf{I})$ by a differentiable sampling function $\Phi(\mathbf{P}_{xy}, g_O(\mathbf{I}))$ (following \cite{saito2019pifu}, we use a bilinear sampling function \cite{jaderberg2015spatial} for $\Phi$). Given the sampled image feature, a function parameterized by another neural network $f_O$ estimates the occupancy of a queried point $\mathbf{P}$ as follows:

\begin{equation}
    O(\mathbf{P}) = f_O(\Phi(\mathbf{P}_{xy}, g_O(\mathbf{I})), P_z) = 
    \begin{cases}
            1 & \text{if $\mathbf{P}$ is inside surface} \\
            0 & \text{otherwise}.
    \end{cases}
\end{equation}

PIFu \cite{saito2019pifu} uses a fully convolutional architecture for $g_O$ to obtain image features that are spatially aligned with the queried 3D point, and a Multilayer Perceptron (MLP) for the function $f_O$, which are trained jointly in an end-to-end manner. Aside from the memory efficiency for high-resolution reconstruction, this representation especially benefits volumetric performance capture, as the spatially aligned image features ensure the 3D reconstruction retains details that are present in input images, \emph{e.g.} wrinkles, hairstyles, and various clothing styles. Instead of $L2$ loss as in \cite{saito2019pifu}, we use a Binary Cross Entropy (BCE) loss for learning the occupancy fields. As it penalizes false negatives and false positives more harshly than the $L2$ loss, we obtain faster convergence when using BCE.

Additionally, the same framework can be applied to texture inference by predicting vector fields instead of occupancy fields as follows:
\begin{equation}
    \mathbf{T}(\mathbf{P}, \mathbf{I}) = f_T(\Phi(\mathbf{P}_{xy}, g_T(\mathbf{I})), \Phi(\mathbf{P}_{xy}, g_O(\mathbf{I})), P_z) = \mathbf{C} \in \mathbb{R}^3,
    \label{eq:pifu_function_2}
\end{equation}
where given a surface point $\mathbf{P}$, the implicit function $\mathbf{T}$ predicts RGB color $\mathbf{C}$. The advantage of this representation is that texture inference can be performed on any surface geometry including occluded regions without requiring a shared 2D parameterization \cite{yamaguchi2018high,lazova3dv2019}. We use the $L1$ loss from the sampled point colors.

Furthermore, we made several modifications to the original implementation of \cite{saito2019pifu} to further improve the accuracy and efficiency. For shape inference, instead of the stacked hourglass \cite{newell2016stacked}, we use HRNetV2-W18-Small-v2 \cite{sun2019deep} as a backbone, which demonstrates superior accuracy with less computation and parameters. We also use conditional batch normalization \cite{dumoulin2016adversarially, de2017modulating, mescheder2018occupancy} to condition the MLPs on the sampled image features instead of the concatenation of these features to the queried depth value, which further improves the accuracy without increasing computational overhead. Additionally, inspired by an ordinal depth regression approach \cite{fu2018deep}, we found that representing depth $P_z$ as a soft one-hot vector more effectively propagates depth information, resulting in faster convergence. For texture inference, we detect the visible surface from the reconstruction and directly use the color from the corresponding pixel, as these regions do not require any inference, further improving the realism of free viewpoint rendering. We provide additional ablation studies to validate our design choices in the appendix.

\paragraph{Inference for Human Reconstruction.}
In \cite{saito2019pifu}, the entire digitization pipeline starts with the dense evaluation of the occupancy fields in 3D, from which the surface mesh is extracted using Marching Cubes \cite{lorensen1987marching}. Then, to obtain the fully textured mesh, the texture inference module is applied to the vertices on the surface mesh. While the implicit shape representation allows us to reconstruct 3D shapes with an arbitrary resolution, the evaluation in the entire 3D space is prohibitively slow, requiring tens of seconds to process a single frame. Thus, acceleration by at least two orders of magnitude is crucial for real-time performance.

\subsection{Real-Time Inference and Rendering}
\label{sec:inference}

\afterpage{
\begin{figure}[!t]
 \centering{
 \includegraphics[width=.9\linewidth]{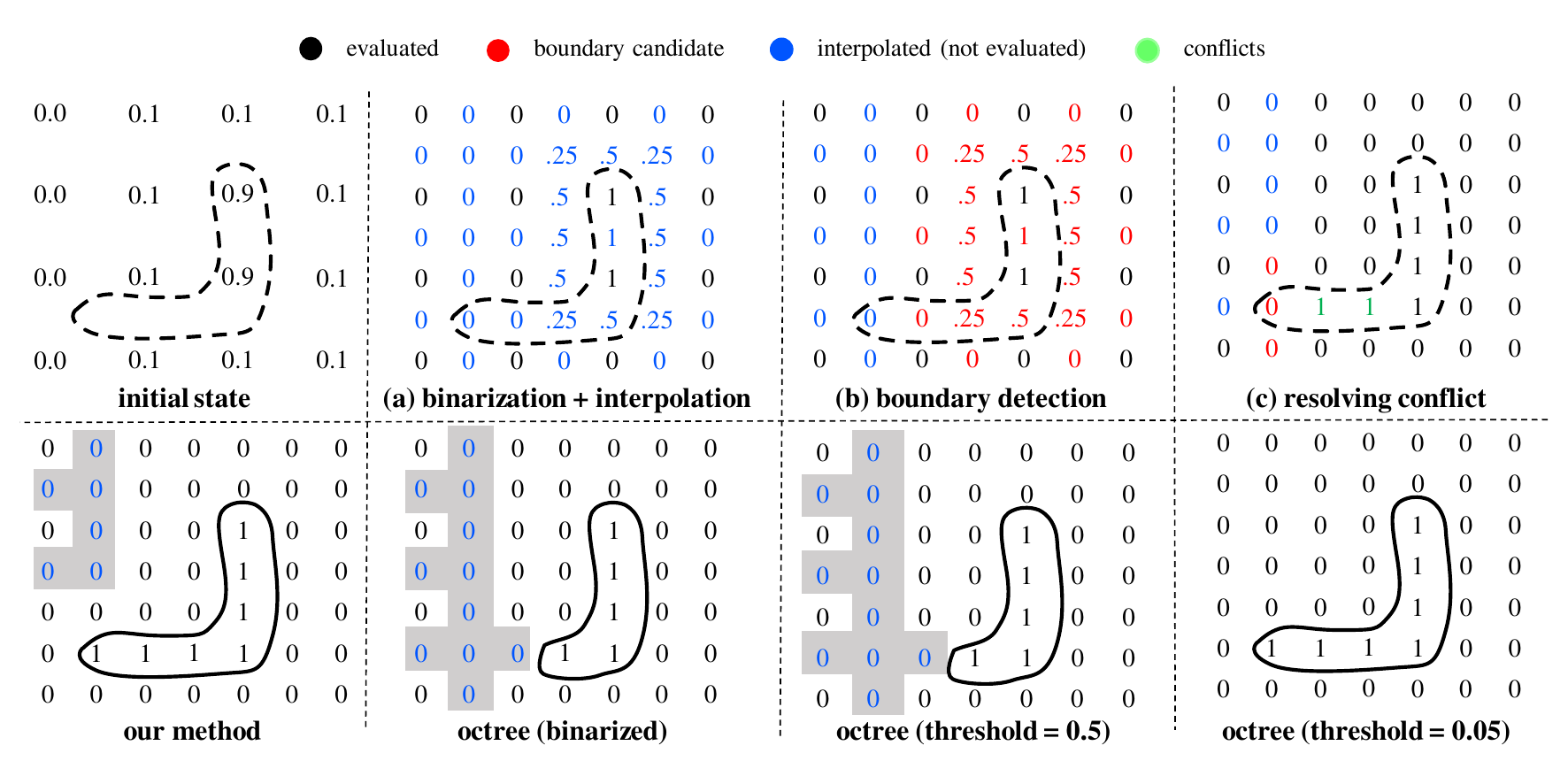}
 \Caption{Our surface localization algorithm overview.}{The dash and solid line denote the true surface and the reconstructed surface respectively. The nodes that are not used for the time-consuming network evaluation are shaded grey.}
 \label{fig:algo_recon}}
\end{figure}
}

\begin{figure*}[!b]
 \centering{
 \includegraphics[trim=0px 0px 0px 0,clip,width=\linewidth]{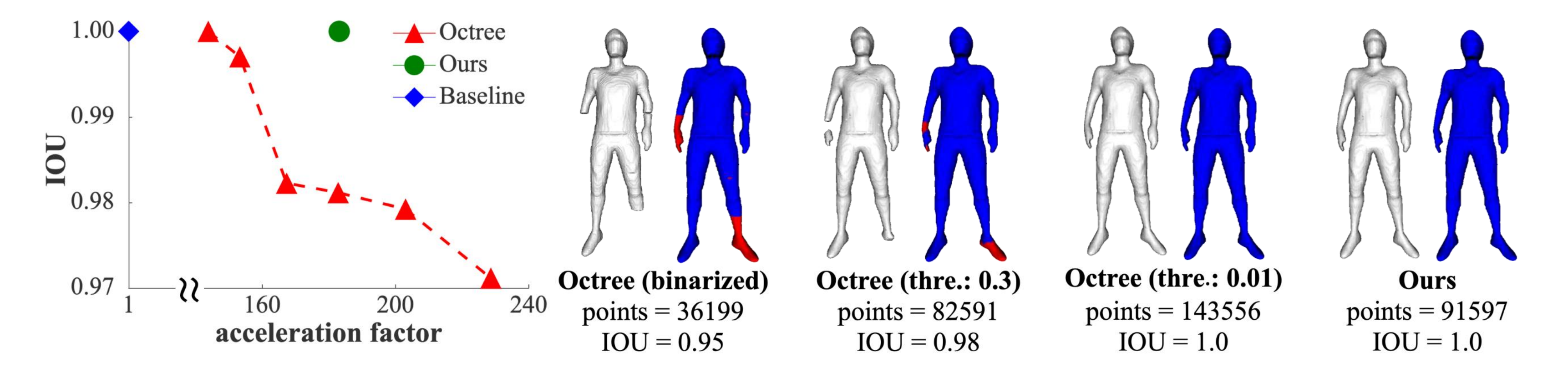}
 \Caption{Comparison of surface reconstruction methods.}{The plot shows the trade-off between the retention of the accuracy of the original reconstruction (\emph{i.e.}, IOU) and speed. The acceleration factor is computed by dividing the number of evaluation points by that with the brute-force baseline. Note that the thresholds used for the octree reconstructions are $0.05$, $0.08$, $0.12$, $0.2$, $0.3$, and $0.4$ from left to right in the plot.}
 \label{fig:algo_comparison}}
\end{figure*}

To reduce the computation required for real-time performance capture, we introduce two novel acceleration techniques. First, we present an efficient surface localization algorithm that retains the accuracy of the brute-force reconstruction with the same complexity as naive octree-based reconstruction algorithms. Furthermore, since our final outputs are renderings from novel viewpoints, we bypass the explicit mesh reconstruction stage by directly generating a novel-view rendering from PIFu. By combining these two algorithms, we can successfully render the performance from arbitrary viewpoints in real-time. We describe each algorithm in detail below. 

\paragraph{Octree-based Robust Surface Localization.}
The major bottleneck of the pipeline is the evaluation of implicit functions represented by an MLP at an excessive number of 3D locations. Thus, substantially reducing the number of points to be evaluated would greatly increase the performance. The octree is a common data representation for efficient shape reconstruction \cite{zhou2010data} which hierarchically reduces the number of nodes in which to store data. To apply an octree for an implicit surface parameterized by a neural network, recently \cite{mescheder2018occupancy} propose an algorithm that subdivides grids only if it is adjacent to the boundary nodes (\emph{i.e.}, the interface between inside node and outside node) after binarizing the predicted occupancy value. We found that this approach often produces inaccurate reconstructions compared to the surface reconstructed by the brute force baseline (see Fig. \ref{fig:algo_comparison}). Since a predicted occupancy value is a continuous value in the range $[0,1]$, indicating the confidence in and proximity to the surface, another approach is to subdivide grids if the maximum absolute deviation of the neighbor coarse grids is larger than a threshold. While this approach allows for control over the trade-off between reconstruction accuracy and acceleration, we also found that this algorithm either excessively evaluates unnecessary points to perform accurate reconstruction or suffers from impaired reconstruction quality in exchange for higher acceleration. To this end, we introduce a surface localization algorithm that hierarchically and precisely determines the boundary nodes. 

We illustrate our surface localization algorithm in Fig. \ref{fig:algo_recon}. Our goal is to locate grid points where the true surface exists within one of the adjacent nodes at the desired resolution, as only the nodes around the surface matter for surface reconstruction. We thus use a coarse-to-fine strategy in which boundary candidate grids are progressively updated by culling unnecessary evaluation points.

Given the occupancy prediction at the coarser level, we first binarize the occupancy values with threshold of $0.5$, and apply interpolation (\emph{i.e.}, bilinear for 2D cases, and trilinear for 3D) to tentatively assign occupancy values to the grid points at the current level (Fig. \ref{fig:algo_recon}(a)). Then, we extract the boundary candidates by extracting the grid points whose values are neither $0$ nor $1$. To cover sufficiently large regions, we apply a dilation operation to incorporate the $1$-ring neighbor of these boundary candidates (Fig. \ref{fig:algo_recon}(b)). These selected nodes are evaluated with the network and the occupancy values at these nodes are updated. Note that if we terminate at this point and move on to the next level, the true boundary candidates may be culled similar to the aforementioned acceleration approaches. Thus, as an additional step, we detect conflict nodes by comparing the binarized values of the interpolation and the network prediction for the boundary candidates. The key observation is that there must be a missing surface region when the value of prediction and the interpolation is inconsistent. The nodes adjacent to the conflict nodes are evaluated with the network iteratively until all the conflicts are resolved (Fig. \ref{fig:algo_recon}(c)). 

Fig. \ref{fig:algo_recon} shows the octree-based reconstruction with binarization \cite{mescheder2018occupancy} and the subdivision with a higher threshold suffers from inaccurate surface localization. While the subdivision approach with a lower threshold can prevent inaccurate reconstruction, an excessive number of nodes are evaluated. On the other hand, our approach not only extracts the accurate surface but also effectively reduces the number of nodes to be evaluated (see the number of blue-colored nodes).

\paragraph{Mesh-free Rendering.}
While the proposed localization algorithm successfully accelerates surface localization, our end goal is rendering from novel viewpoints, and from any viewpoint a large portion of the reconstructed surface is not visible. Furthermore, PIFu allows us to directly infer texture at any point in 3D space, which can substitute the traditional rendering pipeline, where an explicit mesh is required to rasterize the scene. In other words, we can directly generate a novel-view image if the surface location is given from the target viewpoint. Motivated by this observation, we propose a view-based culling algorithm together with a direct rendering method for implicit data representations \cite{saito2019pifu,Oechsle_2019_ICCV,mescheder2018occupancy}. Note that while recently differentiable sphere tracing \cite{liu2019dist} and ray marching \cite{niemeyer2019differentiable} approaches have been proposed to directly render implicit fields, these methods are not suitable for real-time rendering, as they sacrifice computational speed for differentiability to perform image-based supervision tasks.

\begin{figure}[t]
 \centering{
 \includegraphics[width=0.9\linewidth]{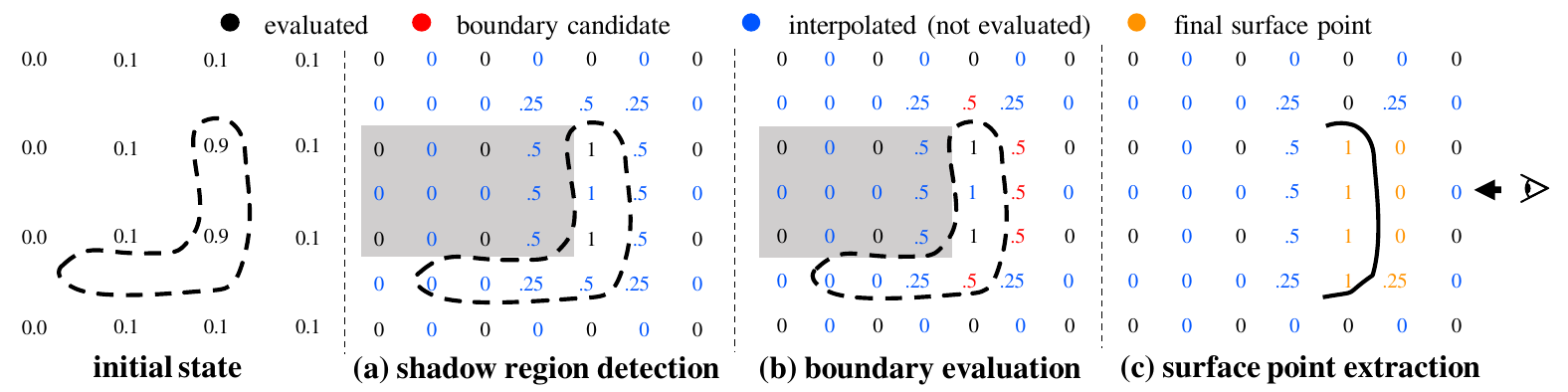}
  \Caption{Our mesh-free rendering overview.}{The dash and solid line denote the true surface and the reconstructed surface respectively. }
 \label{fig:algo_render}}
\end{figure}

Fig. \ref{fig:algo_render} shows the overview of the view-based surface extraction algorithm. For efficient view-based surface extraction, note that the occupancy grids are aligned with the normalized device coordinates defined by the target view instead of the model or world coordinates. That is, the $x$ and $y$ axes in the grid are corresponding to the pixel coordinates and the $z$ axis is aligned with camera rays. Thus, our first objective is to search along the $z$ axis to identify the first two consecutive nodes within which the surface geometry exists. 

First, we apply the aforementioned surface localization algorithm up to the $(L-1)$-th level, where $2^L \times 2^L \times 2^L$ is the target spatial resolution. Then, we upsample the binarized prediction at the $(L-1)$-th level using interpolation and apply the argmax operation along the $z$ axis. The argmax operation provides the maximum value and the corresponding $z$ index along the specified axis, where higher $z$ values are closer to the observer. We denote the maximum value and the corresponding index at a pixel $\mathbf{q}$ by $O_{max}(\mathbf{q})$ and $i_{max}(\mathbf{q})$ respectively. Note that if multiple nodes contain the same maximum value, the function returns the smallest index. If $O_{max}(\mathbf{q}) = 1$, the nodes whose indices are greater than $i_{max}(\mathbf{q})$ are always occluded. Therefore, we treat these nodes as \textit{shadow nodes} which are discarded for the network evaluation (Fig. \ref{fig:algo_render}(a)). Once shadow nodes are marked, we evaluate the remaining nodes with the interpolated value of $0.5$ and update the occupancy values (Fig. \ref{fig:algo_render}(b)). Finally, we apply binarization to the current occupancy values and perform the argmax operation again along the $z$ axis to obtain the updated nearest-point indices. For the pixels with $O_{max}(\mathbf{q}) = 1$, we take the nodes with the index of $i_{max}(\mathbf{q})-1$ and $i_{max}(\mathbf{q})$ as surface points and compute the 3D coordinates of surface $\mathbf{P}(\mathbf{q})$ by interpolating these two nodes by the predicted occupancy value (Fig. \ref{fig:algo_render}(c)). Then a novel-view image $\mathbf{R}$ is rendered as follows:
\begin{equation}
    \mathbf{R}(\mathbf{q}) = 
    \begin{cases}
            \mathbf{T}(\mathbf{P}(\mathbf{q}), \mathbf{I}) & \text{if $O_{max}(\mathbf{q}) = 1$} \\
            \mathbf{B} & \text{otherwise},
        \end{cases}
    \label{eq:pifu_function_1}
\end{equation}
where $\mathbf{B} \in \mathbb{R}^3$ is a background color. For virtual teleportation applications, we composite the rendering and the target scene using a transparent background.

\subsection{Online Hard Example Mining for Data Sampling}
\label{sec:train}

\begin{figure}[t]
 \centering{
 \includegraphics[width=.8\linewidth]{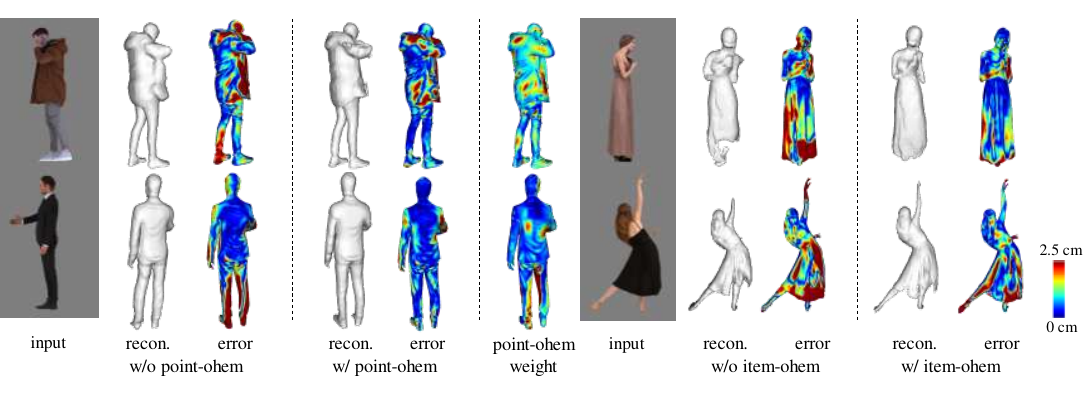}}
 \Caption{Qualitative Evaluation of the OHEM sampling.}{The proposed sampling  effectively selects challenging regions, resulting in significantly more robust reconstruction.}
 \label{fig:ohem_comp}
\end{figure}

As in \cite{saito2019pifu}, the importance-based point sampling for shape learning is more effective than uniform sampling within a bounding box to obtain highly detailed surfaces. However, we observe that this sampling strategy alone still fails to accurately reconstruct challenging poses and viewing angles, which account for only a small portion of the entire training data (see Fig. \ref{fig:ohem_comp}). Although one solution is to synthetically augment the dataset with more challenging training data, manually designing such a data augmentation strategy is non-trivial because various attributes (\emph{e.g.}, poses, view angles, illuminations, and clothing types) may contribute to failure modes, and they are highly entangled.

Nevertheless, the success of importance sampling in \cite{saito2019pifu} illustrates that changing the data sampling distribution directly influences the quality of the reconstruction. 
This observation leads us to a fundamental solution to address the aforementioned training data bias without domain-specifig knowledge. The key idea is to have the network automatically discover hard examples without manual intervention and adaptively change the sampling probability. We will first formulate the problem and solution in a general form and then develop an algorithm for our specific problem. While there are some works address the data bias problem using online hard negative mining (OHEM) strategy in various tasks such as learning image descriptors\cite{simo2014fracking}, image classifiers\cite{loshchilov2015online}, and object detection\cite{shrivastava2016training}, each employs a mining strategy specific to their task.  So it is non-trivial to extend there algorithms to another problem. On the contrary, our formulation is general and can be applied to any problem domain as it requires no domain-specific knowledge.

Given a dataset $\mathcal{M}$, a common approach for supervised learning is to define an objective function $L_m$ per data sample $m$ and reduce an error within a mini-batch using optimizers (\emph{e.g.}, SGD, Adam \cite{kingma2014adam}). Assuming uniform distribution for data sampling, we are minimizing the following function $\mathcal{L}$ w.r.t. variables (\emph{i.e.}, network weights) over the course of iterative optimization:
\begin{equation}
    \mathcal{L} = \frac{1}{\|\mathcal{M}\|}\sum_{m \in \mathcal{M}}\mathcal{L}_m.
    \label{eq:dist1}
\end{equation}
Now suppose the dataset is implicitly clustered into $S$ classes denoted as $\{\mathcal{M}_i\}$ based on various attributes (\emph{e.g.}, poses, illumination). Eq. \ref{eq:dist1} can be written as:
\begin{equation}
    \mathcal{L} = \frac{1}{\|\mathcal{M}\|}\sum_i{(\sum_{m \in \mathcal{M}_i}{\mathcal{L}_m})}
    = \sum_i{P_i \cdot (\frac{1}{\|\mathcal{M}_i\|} \sum_{m \in \mathcal{M}_i}{\mathcal{L}_m})},
    \label{eq:dist2}
\end{equation}
where $P_i = \frac{\|\mathcal{M}_i\|}{\|\mathcal{M}\|}$ is the sampling probability of the cluster $\mathcal{M}_i$ among all the data samples. As shown in Eq. \ref{eq:dist2}, the objective functions in each cluster are weighted by the probability $P_i$. This indicates that hard examples with lower probability are outweighed by the majority of the training data, resulting in poor reconstruction. On the other hand, if we modify the sampling probability of data samples in each cluster to be proportional to the inverse of the class probability $P^{-1}_i$, we can effectively penalize hard examples by removing this bias.

In our problem setting, the goal is to define the sampling probability per target image $P_{\text{im}}$ and per 3D point $P_{\text{pt}}$, or alternatively to define the inverse of these directly. Note that the inverse of probability needs to be positive and not to go to infinity. By assuming the accuracy of prediction is correlated with class probability, we approximate the probability of occurrence of each image by an accuracy measurement as $P_{\text{im}} \sim \text{IoU}$, where IoU is computed by the sampled $n_O$ points for each image. Similarly, we use a Binary Cross Entropy loss to approximate the original probability of sampling points.
Based on these approximations, we model the inverse of the probabilities as follows:
\begin{equation}
    P^{-1}_{\text{im}} = \exp(-\text{IoU}/\alpha_{\text{i}} + \beta_{\text{i}}),\;\;\;\;\;\;
    P^{-1}_{\text{pt}} = \frac{1}{\exp(-\mathcal{L}_{BCE}/\alpha_{\text{p}}) + \beta_{\text{p}}},
    \label{eq:norm_image_points}
\end{equation}
where $\alpha$ and $\beta$ are hyperparameters. In our experiments, we use $\alpha_{\text{i}}=0.15$, $\beta_{\text{i}}=10.0$, $\alpha_{\text{p}}=0.7$ and $\beta_{\text{p}}=0.0$. During training, we compute $P^{-1}_{\text{im}}$ and $P^{-1}_{\text{pt}}$ for each mini-batch and store the values for each data point, which are later used as the online sampling probability of each image and point after normalization. We refer to OHEM for images and points \textit{item-ohem} and \textit{point-ohem} respectively. Please refer to Sec. \ref{sec:evaluation} for the ablation study to validate the effectiveness of our sampling strategy.

\section{Results}
\label{sec:result}

We train our networks using NVIDIA GV100s with $512 \times 512$ images. During inference, we use a Logitech C920 webcam on a desktop system equipped with 62 GB RAM, a 6-core Intel i7-5930K processor, and 2 GV100s. One GPU performs geometry and color inference, while the other performs surface reconstruction, which can be done in parallel in an asynchronized manner when processing multiple frames. The overall latency of our system is on average 0.25 second.

We evaluate our proposed algorithms on the RenderPeople \cite{renderpeople} and BUFF datasets \cite{zhang2017detailed}, and on self-captured performances. In particular, as public datasets of 3D clothed humans in motion are highly limited, we use the BUFF datasets \cite{zhang2017detailed} for quantitative comparison and evaluation and report the average error measured by the Chamfer distance and point-to-surface (P2S) distance from the prediction to the ground truth. We provide implementation details, including the training dataset and real-time segmentation module, in the appendix.

In Fig. \ref{fig:teaser}, we demonstrate our real-time performance capture and rendering from a single RGB camera. Because both the reconstructed geometry and texture inference for unseen regions are plausible, we can obtain novel-view renderings in real-time from a wide range of poses and clothing styles. We provide additional results with various poses, illuminations, viewing angles, and clothing in the appendix and supplemental video.

\begin{table}[t]
\centering{
\resizebox{1.0\linewidth}{!}{%
 \begin{tabular}{l?cccccc?cccc?c}
 \thickhline
 Metric&\multicolumn{2}{c}{Chamfer}&\multicolumn{2}{c}{P2S}&\multicolumn{2}{c?}{Std}&\multicolumn{2}{c}{Chamfer*}&\multicolumn{2}{c?}{P2S*}&\multicolumn{1}{c}{Runtime}\\
&RP&BUFF&RP&BUFF&RP&BUFF&RP&BUFF&RP&BUFF&fps\\
 \thickhline
VIBE\cite{kocabas2019vibe} & - & 5.485 & - & 5.794 & - & 4.279 & - & 10.653 & - & 11.572 & 20\\
 \hline
DeepHuman\cite{Zheng_2019_ICCV} & - & 4.208 & - & 4.340 & - & 4.022 & - & 10.460 & - & 11.389 & 0.0066\\
 \hline
PIFu\cite{saito2019pifu} & 1.684 & 3.629 & 1.743 & 3.601 & 1.953 & 3.744 & 6.796 & 8.417 & 9.127 & 8.552 & 0.033\\
  \thickhline
Ours & 1.561 & 3.615 & 1.624 & 3.613 & 1.624 & 3.631 & 6.456 & 8.675 & 9.556 & 8.934\\
 \hhline{-----------}
  Ours+P & \bf{1.397} & \bf{3.515} & \bf{1.514} & \bf{3.566} & \bf{1.552} & \bf{3.518} & 6.502 & 8.366 & 7.092 & 8.540 & 15\\
 \hhline{-----------}
Ours+P+I & 1.431 & 3.592 & 1.557 & 3.603 & 1.579 & 3.560 & \bf{4.682} & \bf{8.270} & \bf{5.874} & \bf{8.463}\\
 \thickhline
\end{tabular}
}
\caption{Quantitative results. * mean the results of $top$-$k=10$ worst cases. P denotes \textit{point-ohem} and I denotes \textit{item-ohem}.}
\label{tab:tab_eval_err}
}
\end{table}

\subsection{Evaluation}
\label{sec:evaluation}

Fig. \ref{fig:algo_comparison} shows a comparison of surface reconstruction algorithms. The surface localization based on a binarized octree \cite{mescheder2018occupancy} does not guarantee the same reconstruction as the brute-force baseline, potentially losing some body parts. The octree-based reconstruction with a threshold shows the trade-off between performance and accuracy. Our method achieves the best acceleration without any hyperparameters, retaining the original reconstruction accuracy while accelerating surface reconstruction from $30$ seconds to $0.14$ seconds ($7$ fps). By combining it with our mesh-free rendering technique, we require only $0.06$ seconds per frame ($15$ fps) for novel-view rendering at the volumetric resolution of $256^3$, enabling the first real-time volumetric performance capture from a monocular video.

In Tab. \ref{tab:tab_eval_err} and Fig. \ref{fig:ohem_comp}, we evaluate the effectiveness of the proposed Online Hard Example Mining algorithm quantitatively and qualitatively. Using the same training setting, we train our model with and without the \textit{point-ohem} and \textit{item-ohem} sampling. Fig. \ref{fig:ohem_comp} shows the reconstruction results and error maps from the worst $5$ results in the training set. The \textit{point-ohem} successfully improves the fidelity of reconstruction by focusing on the regions with high error (see the \textit{point-ohem} weight in Fig. \ref{fig:ohem_comp}). Similarly, the \textit{item-ohem} automatically supervises more on hard images with less frequent clothing styles or poses, which we expect to capture as accurately as more common poses and clothing styles. As a result, the overall reconstruction quality is significantly improved, compared with the original implementation of \cite{saito2019pifu}, achieving state-of-the-art accuracy (Tab. \ref{tab:tab_eval_err}).

\begin{figure}[!h]
 \centering{
 \includegraphics[width=0.8\linewidth, height=170pt]{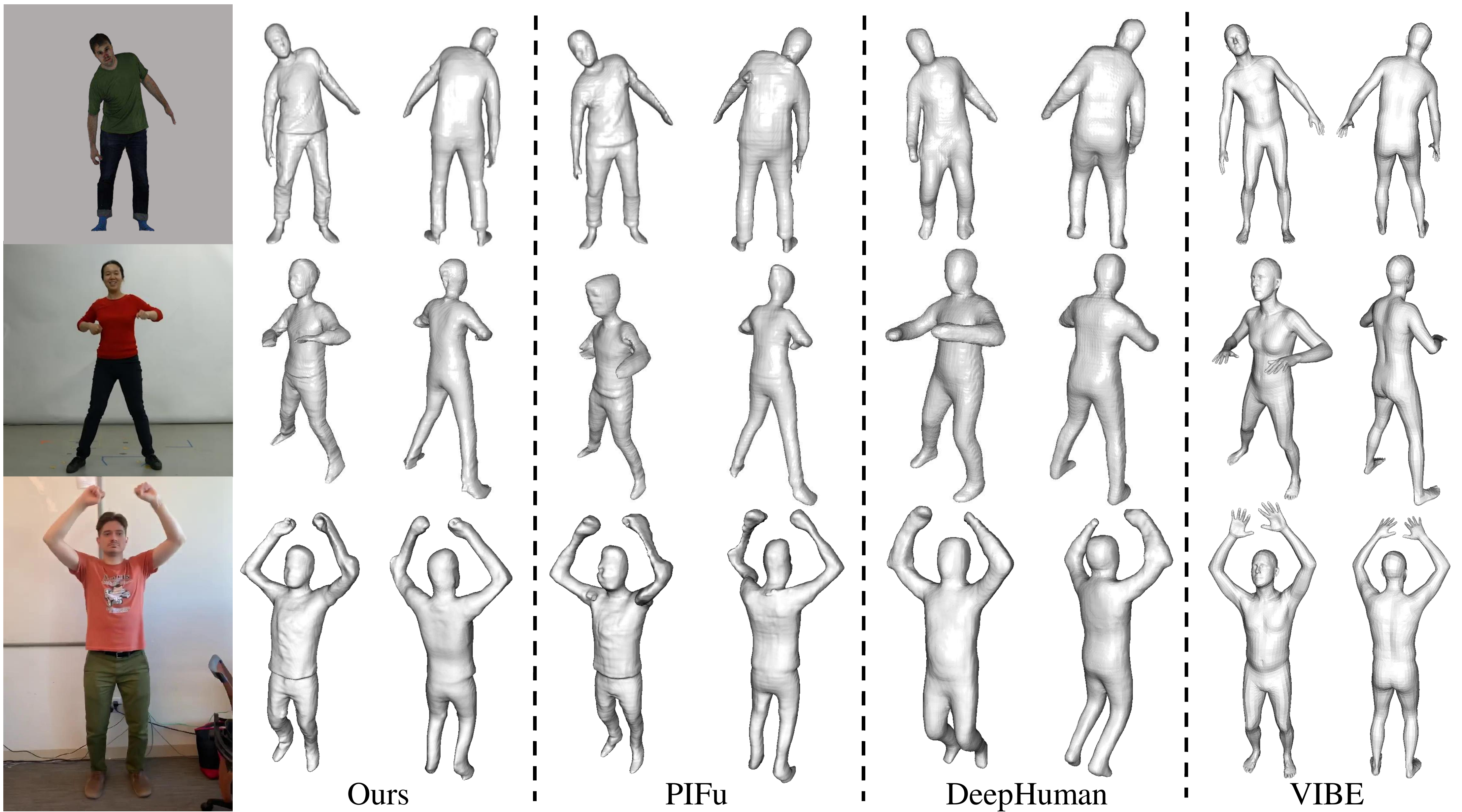}
 \Caption{Qualitative comparison with other reconstruction methods.}{}
 \label{fig:compare_others}}
\end{figure}

\begin{figure}[h]
 \centering{
 \includegraphics[width=0.8\linewidth]{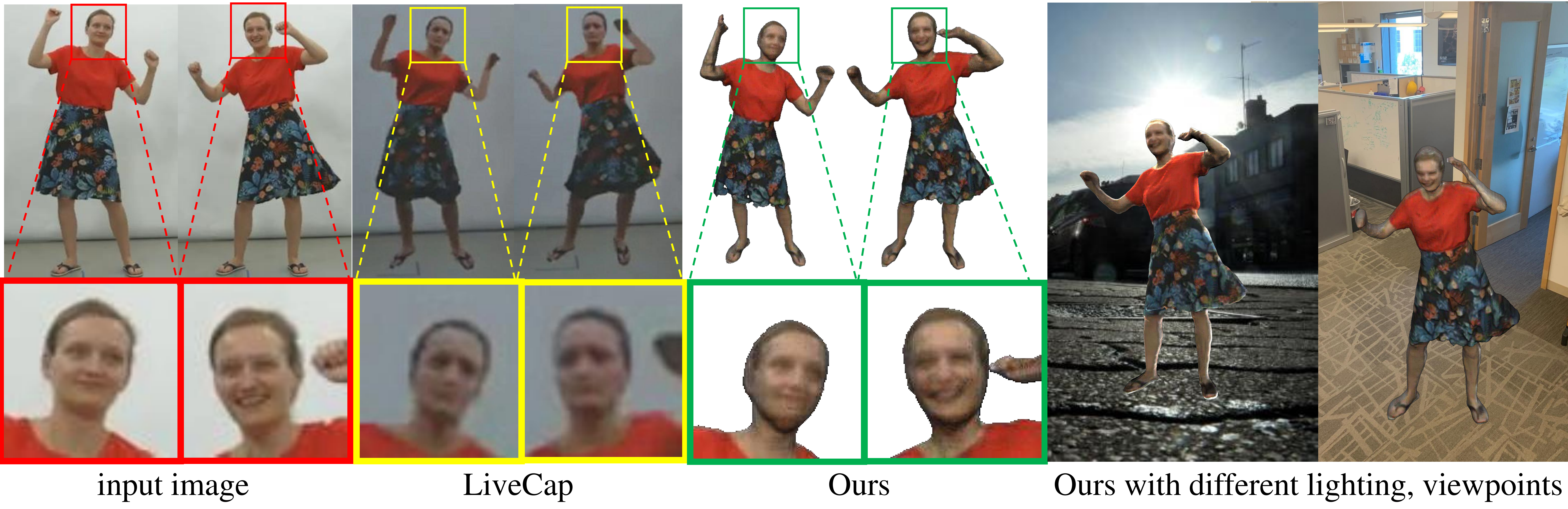}
 \Caption{Comparison with template-based performance capture from monocular video.}{}
 \label{fig:livecap_comp}}
\end{figure}

\subsection{Comparison}

In Tab. \ref{tab:tab_eval_err} and Fig. \ref{fig:compare_others}, we compare our method with the state-of-the-art 3D human reconstruction algorithms from RGB input. Note that we train PIFu \cite{saito2019pifu} using the same training data with the other settings identical to \cite{saito2019pifu} for a fair comparison, while we use the public pretrained models for VIBE \cite{kocabas2019vibe} and DeepHuman \cite{Zheng_2019_ICCV} due to the custom datasets required by each method and their dependency on external modules such as the SMPL \cite{loper2015smpl} model. Although a template-based regression approach \cite{kocabas2019vibe} achieves robust 3D human estimations from images in the wild, the lack of fidelity and details severely impairs the authenticity of the performances. Similarly, a volumetric performance capture based on voxels \cite{Zheng_2019_ICCV} suffers from a lack of fidelity due to the limited resolution. While an implicit shape representation \cite{saito2019pifu} achieves high-resolution reconstruction, the reconstructions become less plausible for infrequent poses and the inference speed (30 seconds) is too slow for real-time applications, both of which we address in this paper. We also qualitatively compare our reconstruction with the state-of-the-art real-time performance capture using a pre-captured template \cite{habermann2019livecap} (Fig. \ref{fig:livecap_comp}). While the reconstructed geometries are comparable, our method can render performances with dynamic textures that reflect lively expressions, unlike a tracking method using a fixed template. Our approach is also agnostic to topology changes, and can thus handle very challenging scenarios such as changing clothing (Fig. \ref{fig:teaser}).

\section{Conclusion}
\label{sec:conclusion}

We have demonstrated that volumetric reconstruction and rendering of humans from a single input image is possible to achieve in near real-time speed without sacrificing the final image quality. 
Our novel progressive surface localization method allows us to vastly reduce the number of points queried during surface reconstruction, giving us a speedup of two orders of magnitude without reducing the final surface quality.
Furthermore, we demonstrate that directly rendering novel viewpoints of the captured subject is possible without explicitly extracting a mesh or performing naive, computationally intensive volumetric rendering, allowing us to obtain real-time rendering performance with the reconstructed surface.
Finally, our Online Hard Example Mining technique allows us to find and learn the appropriate response to challenging input examples, thereby making it feasible to train our networks with a tractable amount of data while attaining high-quality results with large appearance and motion variations.

While we demonstrate our approach on human subjects and performances, our acceleration techniques are straightforward to implement and generalize to any object or topology.
We thus believe this will be a critical building block to virtually teleport anything captured by a commodity camera anywhere.

\section{Acknowledgement}
This research was funded by in part by the ONR YIP grant N00014-17-S-FO14, the CONIX Research Center, a Semiconductor Research Corporation (SRC) program sponsored by DARPA, the Andrew and Erna Viterbi Early Career Chair, the U.S. Army Research Laboratory (ARL) under contract number W911NF-14-D-0005, Adobe, and Sony.

\clearpage
\bibliographystyle{splncs04}
\bibliography{paper}

\clearpage

\renewcommand{\thesection}{A\arabic{section}}
\setcounter{section}{0}

\setlength{\textfloatsep}{0.1cm}

\section{Implementation Details}
\subsection{Datasets}

\paragraph{\textbf{RenderPeople}}
Similar to \cite{saito2019pifu}, we leverage high-quality photogrammetry scans of clothed humans with synthetic rendering to construct our training dataset. Aside from $466$ static scans from RenderPeople\cite{renderpeople} used in \cite{saito2019pifu}, we incorporate additional $167$ rigged models from RenderPeople \cite{renderpeople} and apply $32$ animation sets from Mixamo \cite{mixamo} so that wider pose variations are covered for performance capture. Please refer to appendix \ref{appendix:maximo} for the complete list of animations. By randomly selecting $3$ frames from each animation, we obtain $466 + 167 \times 32 \times 3 = 16,498$ models. We split this into \emph{training} and \emph{validation} sets based on subject identities, resulting in $452 + 164 \times 32 \times 3 = 16,196$ meshes in the \emph{training} set and $14 + 3 \times 32 \times 3 = 302$ meshes in the \emph{validation} set. For training, each mesh is rendered with weak perspective camera at every $10$ degrees around the yaw axis using Precomputed Radiance Transfer \cite{sloan2002precomputed} and $163$ second-order spherical harmonics derived from HDRI Haven \cite{hdrihaven}. For validation, we compute our loss metrics on the \emph{validation} set rendered with $3$ views sampled at $120$-degree intervals around the yaw axis. The hyper-parameters $\alpha_{\text{i}}$, $\beta_{\text{i}}$, $\alpha_{\text{p}}$ and $\beta_{\text{p}}$ in our Online Hard Example Mining (OHEM) training strategy (see Eq. 6 in the paper) are chosen using the \emph{validation} set.

\begin{figure}[t]
 \centering{
 \includegraphics[width=.9\linewidth]{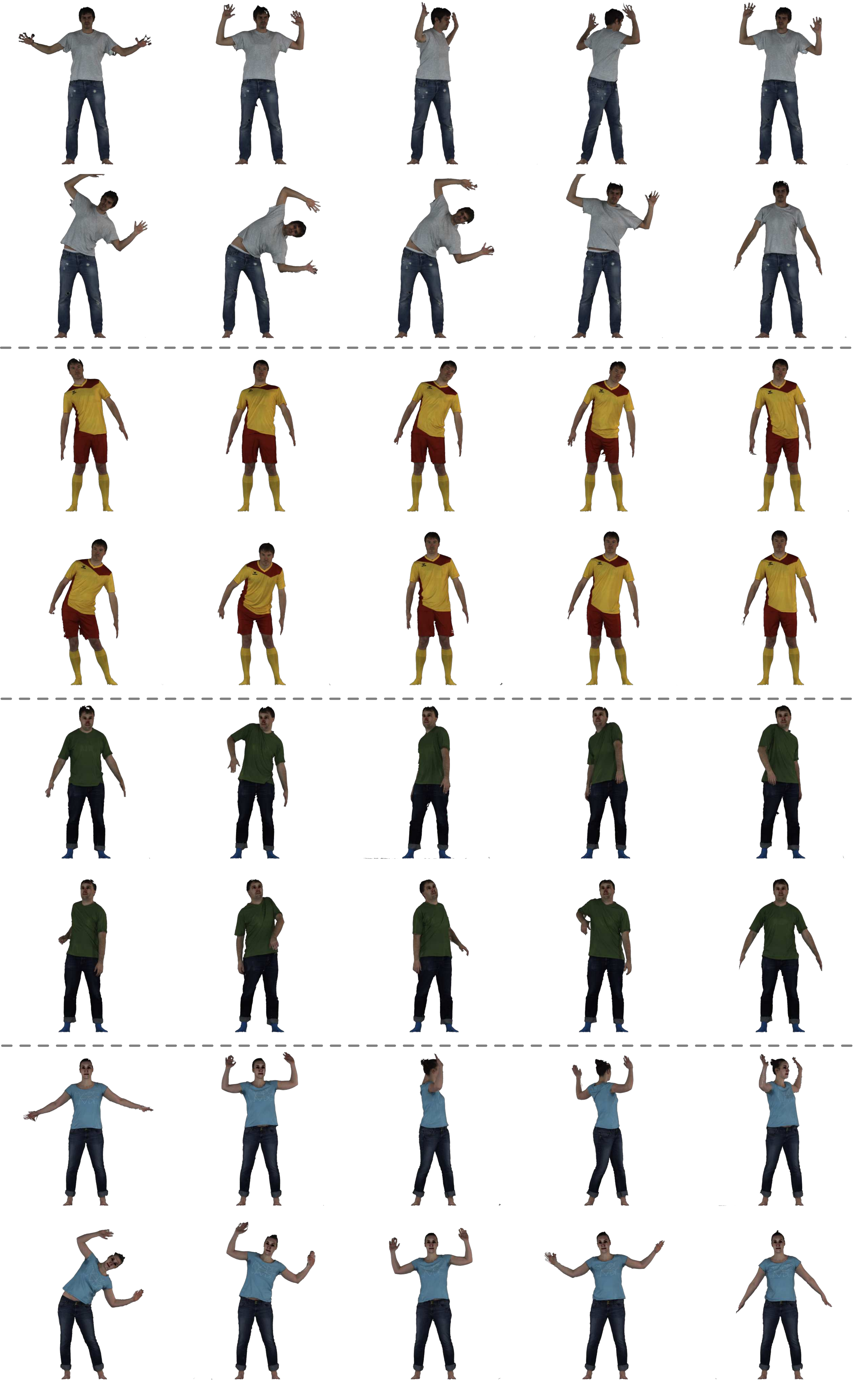}}
 \Caption{Sampled BUFF benchmark.}{We apply K-Medoids to each sequence of BUFF dataset to construct the \emph{test} set. Sufficient pose variations in BUFF dataset are covered with $K=10$.}
 \label{fig:suppl_buff}
\end{figure}

\paragraph{\textbf{BUFF}}
To quantitatively evaluate the generalization ability of the proposed system and fairly compare with the existing methods, we propose to use the BUFF dataset \cite{zhang2017detailed} for the following reasons: First, the BUFF dataset provides high-fidelity geometry with photorealistic texture, approximating the modality of real images with detailed ground truth geometry. Secondly it contains large pose variations. Thus, accuracy of each method under various poses can be properly evaluated. Lastly, as the existing approaches \cite{kocabas2019vibe,Zheng_2019_ICCV,saito2019pifu} are trained with custom datasets, we can fairly compare on a dataset with which none of these methods are trained. The BUFF dataset consists of $5$ subjects, each of which is captured with $1$ or $2$ unique outfits. In total, it contains $26$ sequences with per-frame ground-truth 3D meshes and textures. As the large portion of poses are duplicated (e.g., T-pose), we apply K-Medoids to each sequence to obtain distinctive frames. By setting $K=10$, we obtain $26 \times 10 = 260$ frames and render them from $3$ views points at $120$-degree intervals around the yaw axis, resulting in $260 \times 3 = 780$ images for \emph{test} set (see Fig.~\ref{fig:suppl_buff} for sample images).

\subsection{Network Architectures}
\begin{figure}[t]
 \centering{
 \includegraphics[width=\linewidth]{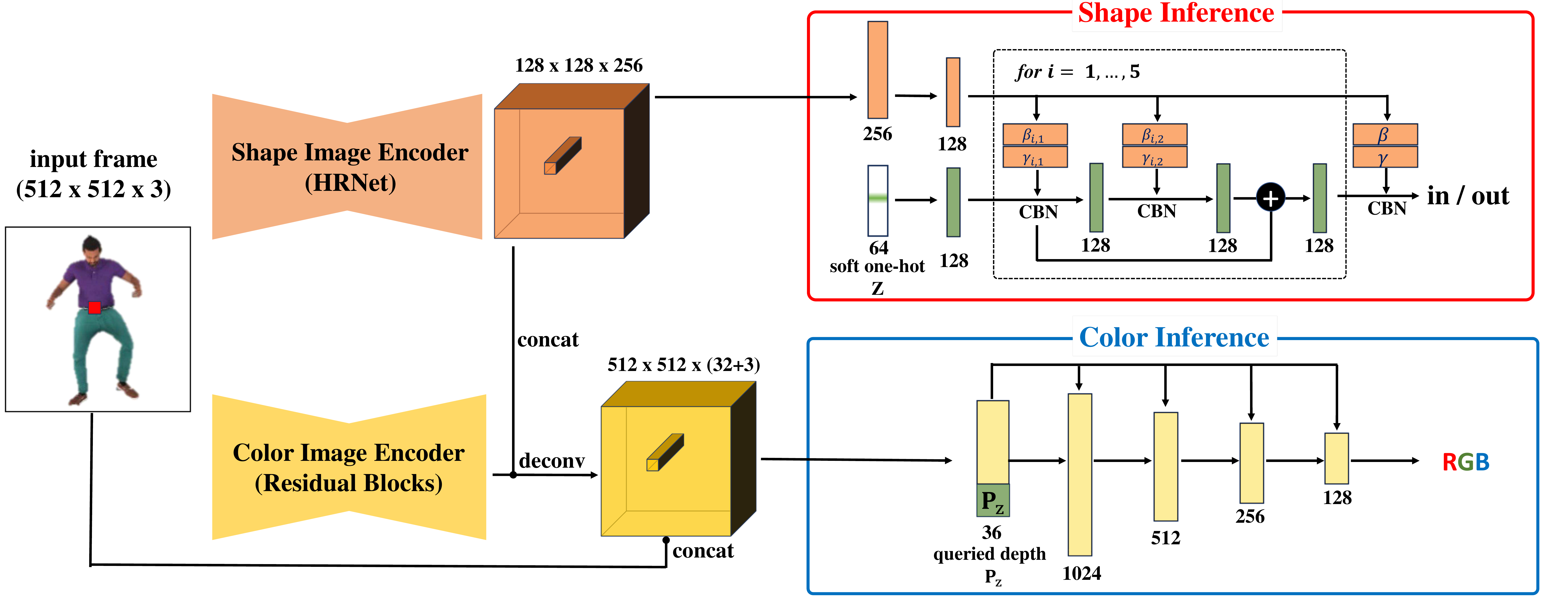}
 \Caption{The overview of our network architectures.}{}
 \label{fig:overview_training}}
\end{figure}

We have made several architectural modifications to improve the efficiency and robustness of the original implementation of \cite{saito2019pifu}. In this section, we provide the implementation details as well as discussion about the effectiveness of each modification. Fig. \ref{fig:overview_training} shows the overview of our network architectures.

\paragraph{Image Encoder} 
For surface reconstruction, we replace the stacked hourglass network~\cite{newell2016stacked} with HRNetV2-W18-Small-v2~\cite{sun2019deep} in our image encoder for shape inference due to its superior performance in various tasks (e.g., semantic segmentations, human pose estimations) with faster computation (see Figure~\ref{fig:overview_training}). The final feature resolution is $128\times128$ with the channel size of $256$ as in~\cite{saito2019pifu}. Table~\ref{table:suppl_abla} shows the ablation study on the choice of image encoders. HRNet not only shows better reconstruction accuracy but also faster runtime ($14$ fps vs $12$ fps) with less parameters and computation. For color inference, we found that a higher spatial resolution for image features result in more detailed textures. To this end, we modify the architecture with $6$ residual blocks \cite{johnson2016perceptual} by upsampling the stacked output feature maps from shape and color image encoders from $128\times128$ to $512\times512$ with the output channel size of $32$ using a transposed convolution.

\begin{figure}[!h]
 \centering{
 \includegraphics[width=0.8\linewidth]{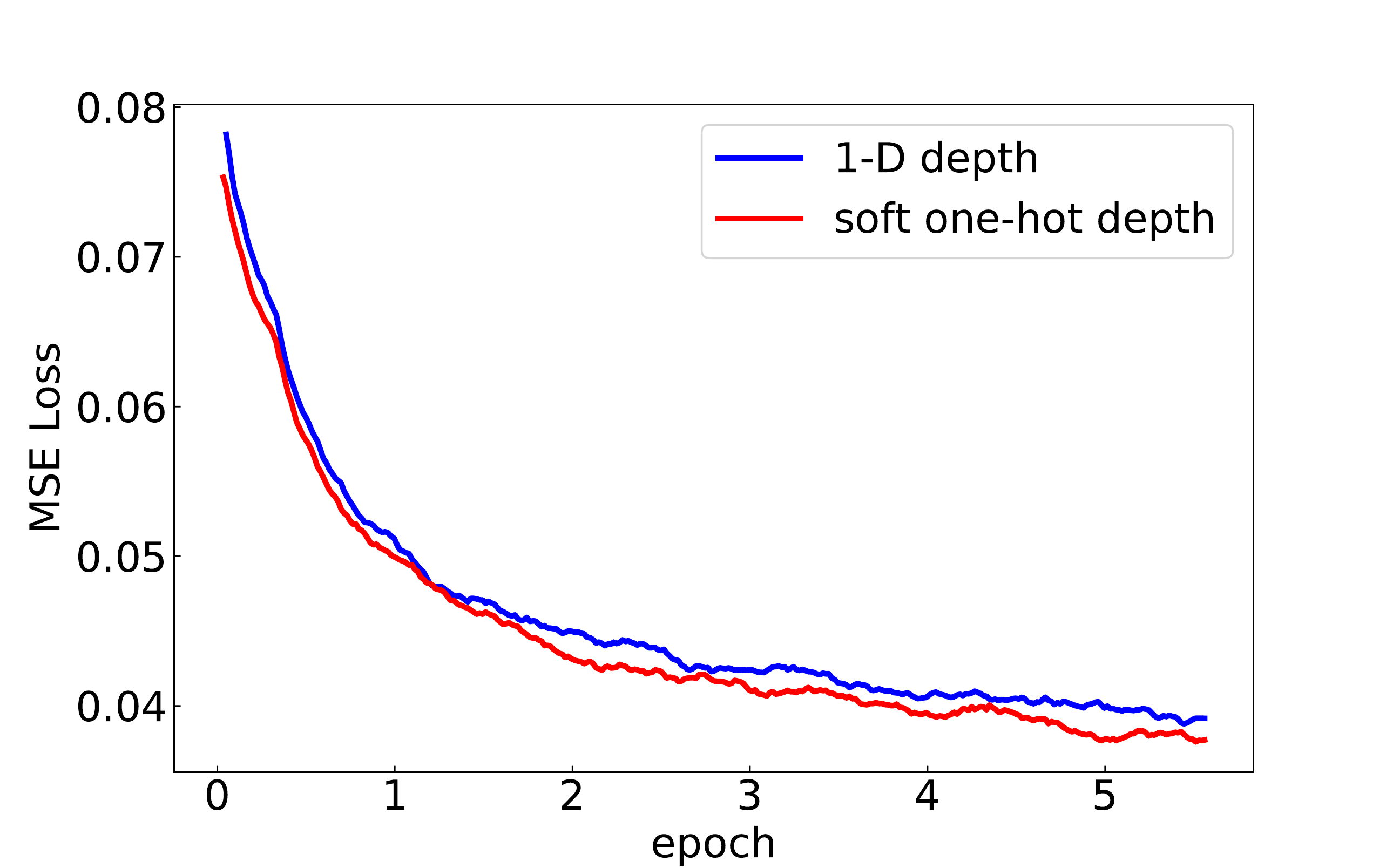}}
 \Caption{Comparison on different depth representation.}{Representing depth $z$ as a \emph{soft} one-hot vector makes the network converge faster with higher accuracy. We use HRNet with CBN for both 1-D depth representation baseline and SoftZ.}
 \label{fig:suppl_ztrick}
\end{figure}

\paragraph{Depth Representation}
Additionally, inspired by a multi-channel depth representation used in ordinal depth regression \cite{fu2018deep},  we found that representing depth $P_z$ as a multi-dimensional vector more effectively propagates depth information to the shape inference function $f_O$. More specifically, we convert $\{P_z \in \mathbb{R} | -1 \leq P_z \leq 1 \}$ into a $N$-dimensional feature $\mathbf{Z}=\{Z_i\}_{i=0}^{N-1}$ as follows:
\begin{equation}
    Z_i = 
    \begin{cases}
            1 + \left \lfloor{(N-1) \cdot P'_z}\right \rfloor - (N-1) \cdot P'_z & \text{if $i = \left \lfloor{(N-1) \cdot P'_z}\right \rfloor$} \\
            (N-1) \cdot P'_z - \left \lfloor{(N-1) \cdot P'_z}\right \rfloor & \text{if $i = \left \lfloor{(N-1) \cdot P'_z}\right \rfloor + 1$} \\
            0 & \text{otherwise},
    \end{cases}
    \label{eq:pifu_z_encode}
\end{equation}
where $P'_z = 0.5 \cdot (P_z + 1.0)$ and $N=64$ in our experiments. We term this multi-channel depth representation soft one-hot depth (SoftZ). Figure~\ref{fig:suppl_ztrick} and Table~\ref{table:suppl_abla} demonstrates the faster convergence and more accurate reconstruction of the proposed depth representation. 

\paragraph{Pixel-aligned 3D Lifting}
The original implementation of \cite{saito2019pifu} lifts the pixel-aligned image features into 3D by feeding the image feature and the depth value $P_z$ into a multi-layer perceptron (MLP). To further reduce the channel size of intermediate layers, we adopt a conditional batch normalization (CBN)~\cite{dumoulin2016adversarially, de2017modulating,mescheder2018occupancy}. More specifically, the soft one-hot depth vector $\mathbf{Z}$ (our final model) or the depth value $P_z$ (only for ablation study) is fed into a multi-layer perceptron (MLP) consisting of $5$ blocks of an conditional batch normalization module (CBN)~\cite{dumoulin2016adversarially, de2017modulating,mescheder2018occupancy} where input feature vector for each CBN layer are normalized with the learnable multiplier $\gamma(c)$ and bias $\beta(c)$ taking as input a conditional vector $c$ as follows:
\begin{equation}
    f_{out} = \gamma(c)\frac{f_{in}-\mu}{\sqrt{\sigma^2+\epsilon}} + \beta(c),
\end{equation}
where $f_{in}$ and $f_{out}$ are the input and output features, $\mu$ is the statistical mean, $\sigma$ is the standard deviation, and $\epsilon=1.0\times10^{-5}$. Each layer is followed by non-linear ReLU activation. Note that unlike \cite{mescheder2018occupancy}, our conditional vector $c$ is pixel-aligned image features $\Phi(\mathbf{P}_{xy}, g_O(\mathbf{I}))$ to learn precise geometry aligned with an input image. Please refer to Figure~\ref{fig:overview_training} for the detailed architecture. We use the channel size of $128$ for all the intermediate feature dimensions. In Tab. \ref{table:suppl_abla}, CBN is referred to as 3d lifting with conditional batch normalization modules and MLP as the naive concatenation of a queried depth value and image features as in \cite{saito2019pifu}. The number of parameters and computational overhead are further reduced while retaining the same level of reconstruction accuracy.

For color inference, we take as input the concatenation of the depth value $P_z$, RGB value from the corresponding pixel of the input image, and the learned image feature, resulting in $36$ dimensional vector. They are fed into another MLP consisting of $5$ layers with the channel size of $1024$, $512$, $256$, $128$, and $3$ and skip connections at $1$, $2$, $3$, and $4$-th layers. Each layer is followed by the LeakyReLU activation except the last layer, and Tanh activation for the last layer.

\paragraph{Training procedure}
We use RMSProp~\cite{Tieleman2012} and Adam~\cite{kingma2014adam} for the surface reconstruction and texture inference respectively, with a learning rate of $1e-3$. Since the batch normalization layer in HRNet and CBN can benefit from large batch sizes, we use a batch size of $24$ for both surface reconstruction and texture inference. The number of sampled points per image is $4096$ in every training batch. We first train the surface reconstruction network for $5$ epochs with the constant learning rate, then fix it and only train the texture inference network for $5$ more epochs. The training of our networks for surface reconstruction and texture inference takes $3$ days each on a single NVIDIA GV100 GPU.

\subsection{Real-time Human Segmentation}
\begin{figure}[t]
 \centering{
 \includegraphics[width=.9\linewidth]{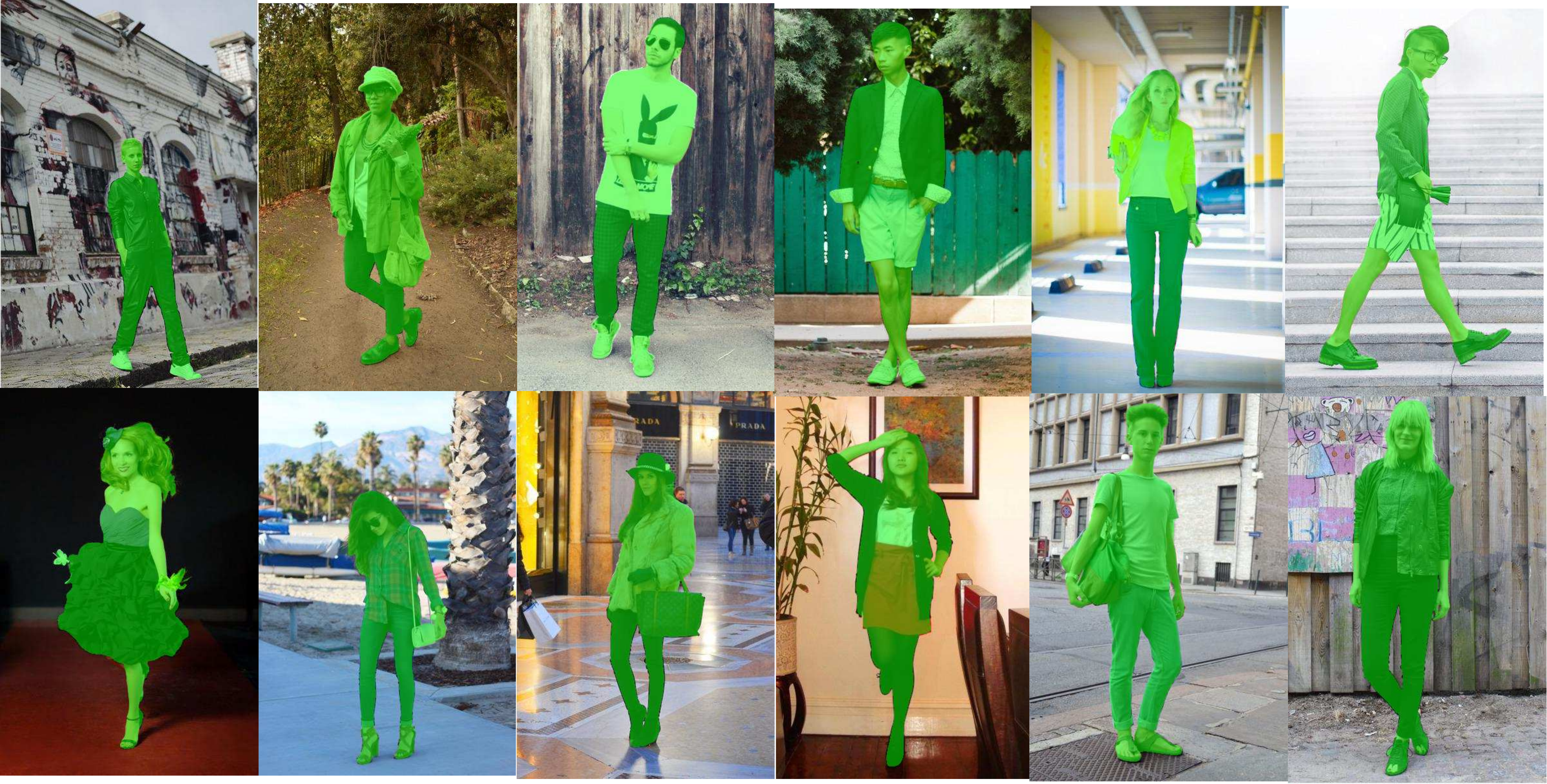}}
 \Caption{Training data for our real-time segmentation network.}{}
 \label{fig:suppl_seg_data}
\end{figure}

\begin{figure}[t]
 \centering{
 \includegraphics[width=.9\linewidth]{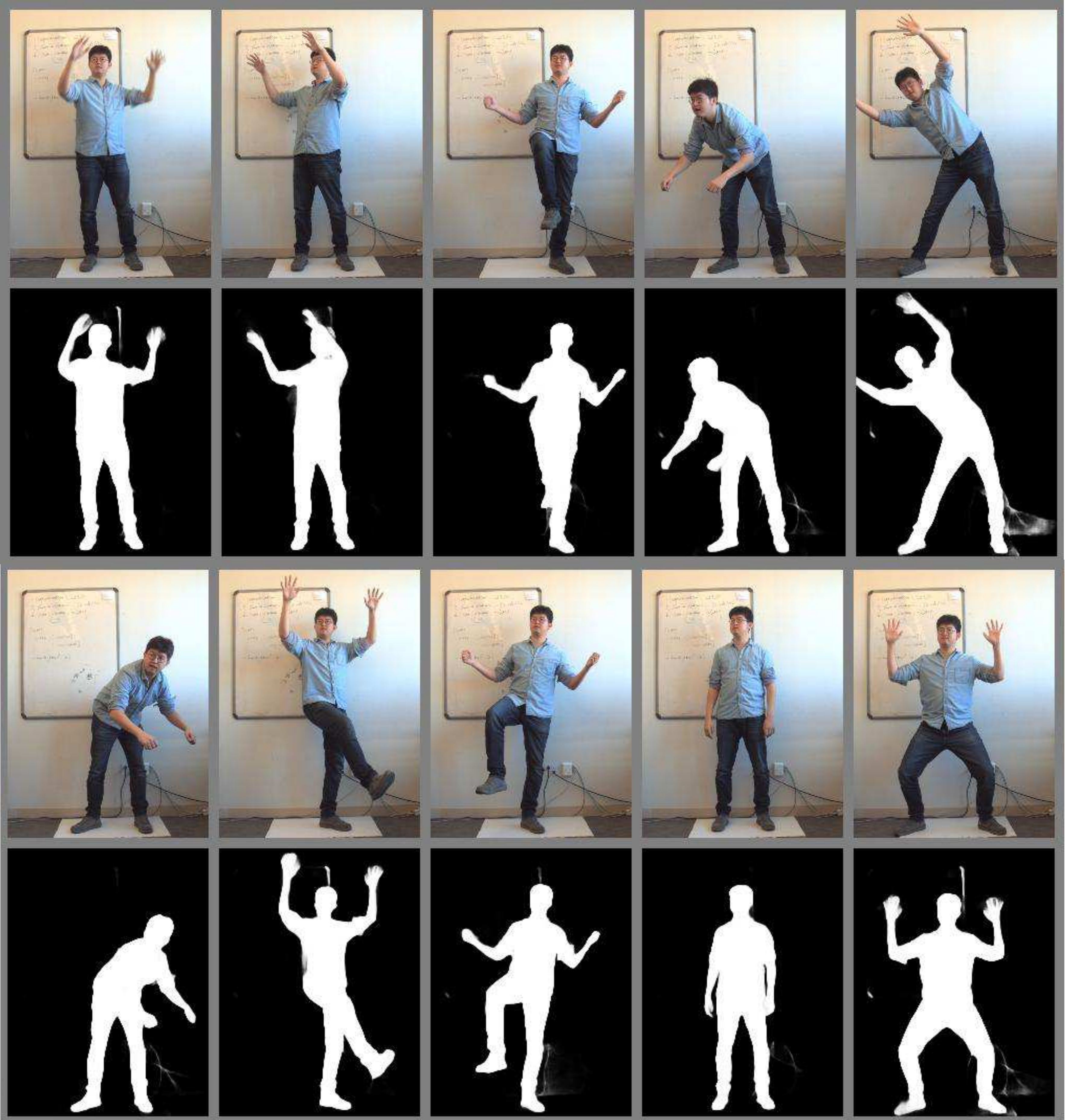}}
 \Caption{Results of our segmentation network.}{}
 \label{fig:suppl_seg_results}
\end{figure}

As preprocessing, we require an efficient and accurate human segmentation network. To this end, we start by collecting high-quality data with accurate annotations. Because publicly available human segmentation datasets are either low-quality or biased to particular types of images (\emph{e.g.}, portraits)~\cite{lin2014coco,shen2016automatic,zhang2019pose2seg,gong2017look}, we collected $12,029$ human images with various backgrounds, lighting conditions, poses, and different outfits. Most of the images come from the LIP dataset~\cite{gong2017look}, while the rest are collected from the internet. We obtained high-quality annotations of these images using a commercial website\footnote{\url{https://www.remove.bg/}}. We use a U-Net~\cite{ronneberger2015u} with ResNet-18~\cite{he2016deep} as backbone with Adadelta~\cite{zeiler2012adadelta} using an initial learning rate of $10.0$. The learning rate is reduced by a factor of $0.95$ after each epoch. The training converges after $100$ epoches, which takes about $2$ days on a single NVidia GV100. During inference, with $256\times256$ image resolution, this model run at $150$ fps on NVidia GV100. Figure~\ref{fig:suppl_seg_data} and Figure~\ref{fig:suppl_seg_results} show the sampled training dataset and segmentation results of our real-time segmentation model, respectively.

\begin{table}[t]
\label{table:suppl_abla}
\centering{
\resizebox{1.0\linewidth}{!}{%
 \begin{tabular}{l?cccc?ccc}
 \thickhline
 Metric&\multicolumn{2}{c}{Chamfer}&\multicolumn{2}{c?}{P2S}&Params&GFLOPS&Runtime\\
&RP&BUFF&RP&BUFF& (M) & (per 4096 calls)&(fps)\\
\thickhline
HG-MLP\cite{saito2019pifu} & 1.684 & 3.629 & 1.743 & 3.601 & 15.6 & 105.0+9.7 & 12 \\
\hline
HRNet-MLP & 1.602 & 3.623 & 1.691 & 3.617 & 8.8 & 16.0+9.7 & 14 \\
\hline
HRNet-CBN & 1.584 & 3.626 & 1.652 & \textbf{3.585} & 8.3 & 16.0+3.0 & 15 \\
\thickhline
HRNet-CBN-SoftZ & \textbf{1.561} & \textbf{3.615} & \textbf{1.624} & 3.613 & 8.3 & 16.0+3.0 &15 \\
\thickhline
\end{tabular}
}
}
\caption{Ablation study.}
\end{table}

\section{Additional Results}
\begin{figure}[!h]
 \centering{
 \includegraphics[width=\linewidth]{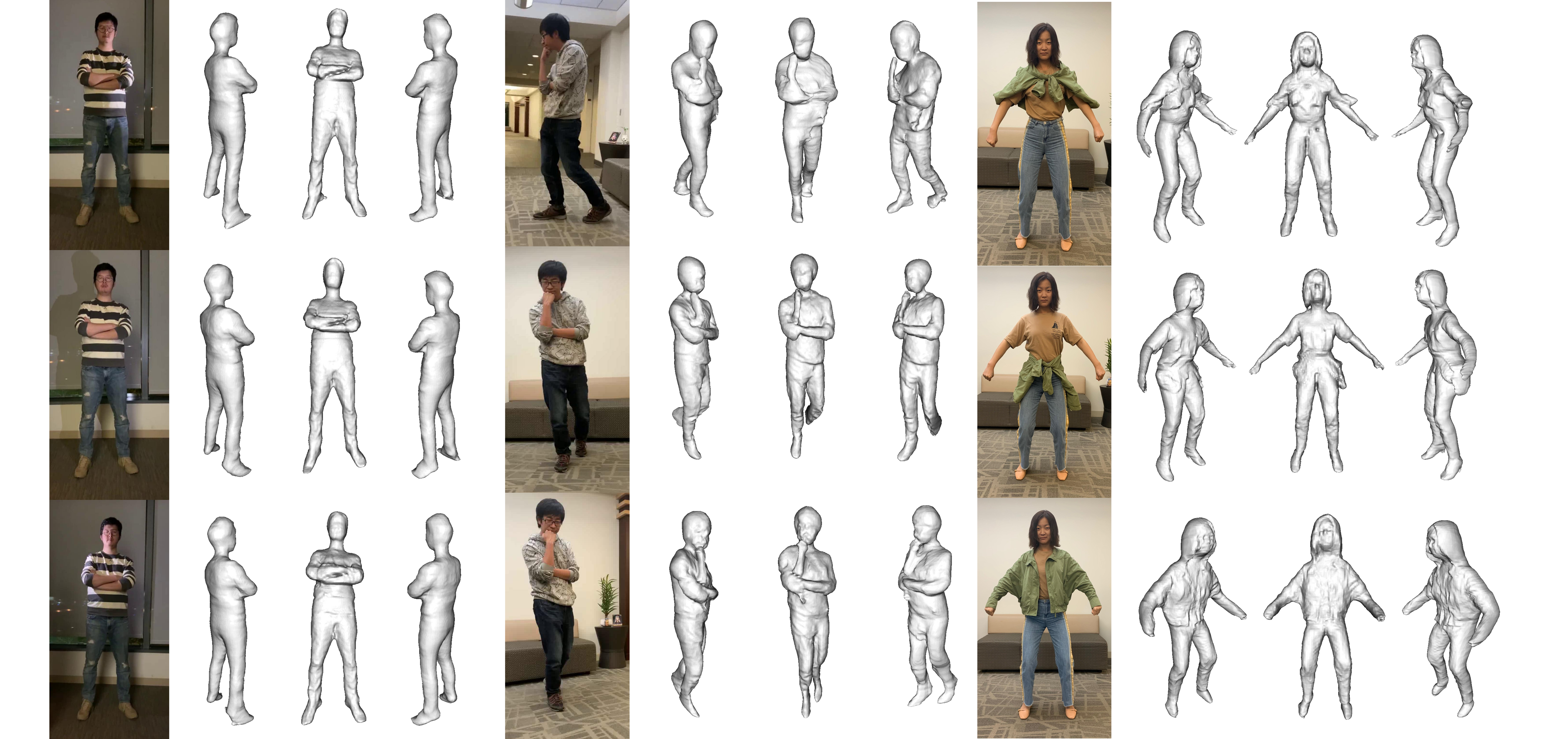}
 \Caption{We qualitatively evaluate the robustness of our approach by demonstrating the consistency of reconstruction with different lighting conditions, viewpoints and surface topology.}{}}
 \label{fig:ablation_light}
\end{figure}

\begin{figure}[t]
 \centering{
 \includegraphics[width=1.0\linewidth]{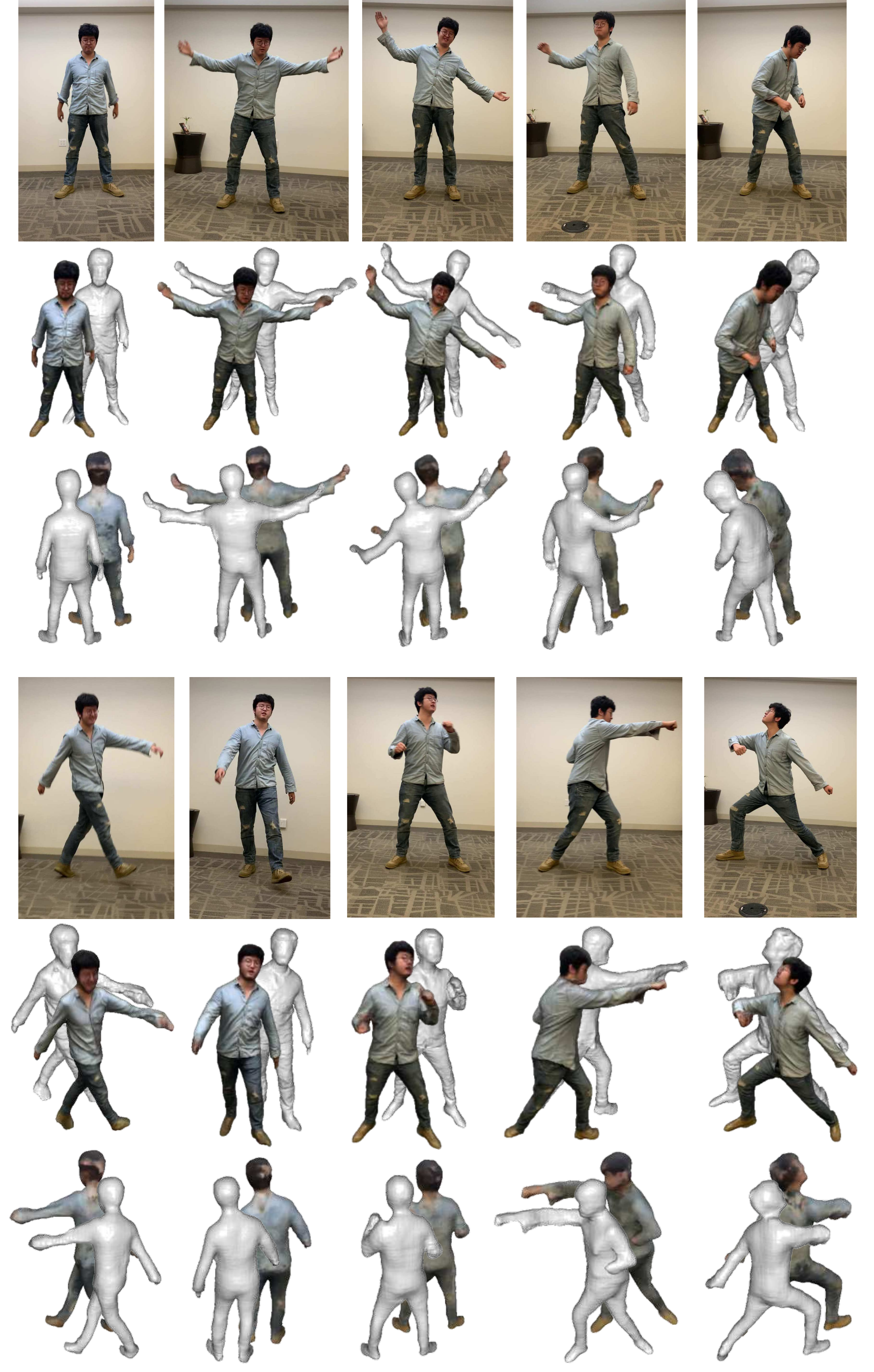}}
 \Caption{Qualitative results on self-captured performances.}{}
 \label{fig:suppl_results1}
\end{figure}

We evaluate the robustness of our algorithm under different lighting conditions, viewpoints, and clothes topology in Figure~\ref{fig:ablation_light}. We also provide additional qualitative results from a video sequence (see Figure~\ref{fig:suppl_results1}) and from internet photos (see Figure~\ref{fig:suppl_results2}). The other video reconstruction results can be found in the supplemental video.

\subsection{Limitations}
\begin{figure}[!h]
 \centering{
 \includegraphics[width=\linewidth]{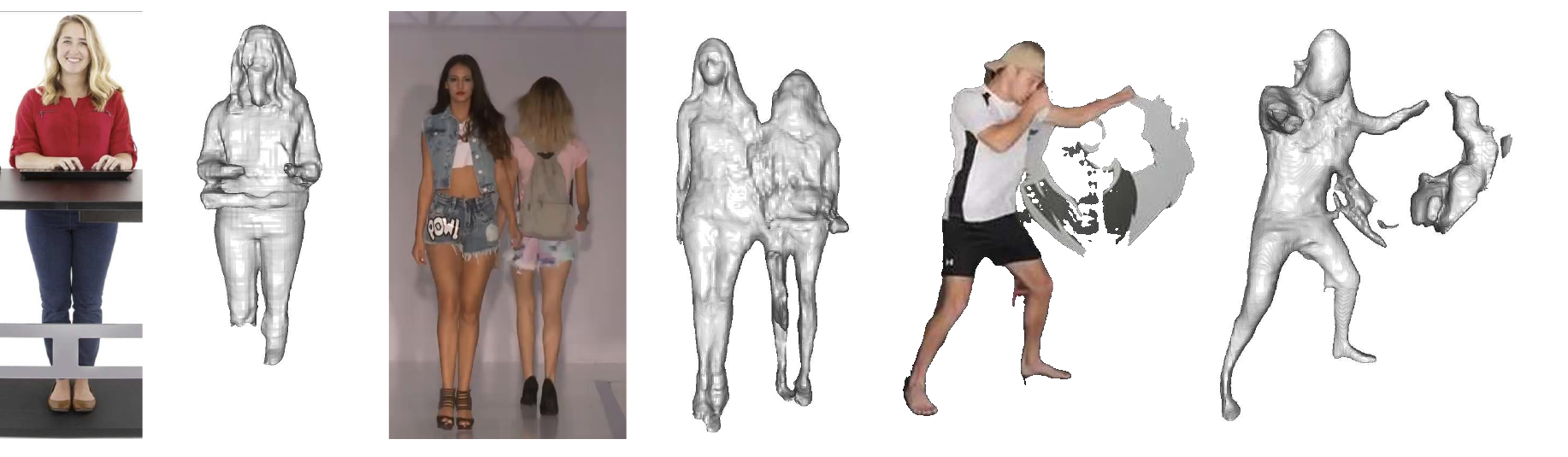}
 \caption{Limitations.}{Our current system may fail in the presence of inaccurate segmentation, multiple subjects, and severe occlusions.}
 \label{fig:limitation}}
\end{figure}

As our training data consists of only a single person at a time, the presence of multiple people confuses the network (see Figure~\ref{fig:limitation}). Modeling multiple subjects \cite{liu2011markerless,joo2018total} is essential to understanding social interaction for a truly believable virtual experience. In the future, we plan to extend our approach to handle multiple people in a single monocular video. Another interesting direction is to handle occlusion by other objects, as a complete 3D reconstruction is difficult without explicitly modeling the occlusion occurring in natural scenes.

\begin{figure}[t]
 \centering{
 \includegraphics[width=1.0\linewidth]{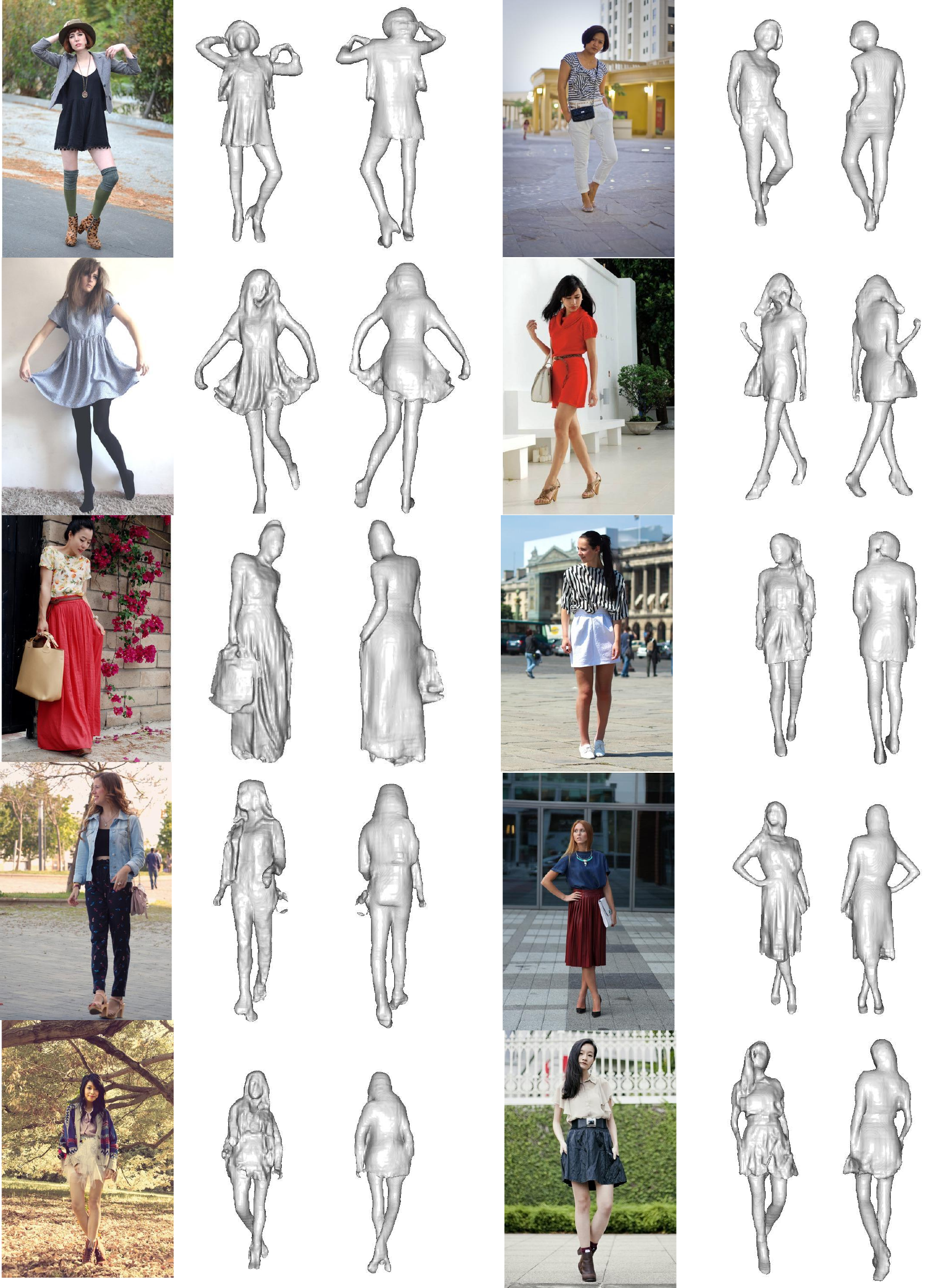}}
 \Caption{Qualitative results on internet photos.}{}
 \label{fig:suppl_results2}
\end{figure} 

\clearpage

\appendix
\section{Mixamo Animation Sets}{
\label{appendix:maximo}
\begin{center}
\begin{tabular}{ |c |c|}
\hline
Agreeing & Bored \\
Breakdance\_Ready & Defeat  \\
Defeated & Dwarf\_Idle  \\
Female\_Tough\_Walk & Hands\_Forward\_Gesture  \\
Holding\_Idle & Look\_Over\_Shoulder  \\
Old\_Man\_Idle & Orc\_Idle \\
Patting & Pointing \\
Put\_Back\_Rifle\_Behind\_Shoulder & Searching\_Files\_High \\
Shoulder\_Rubbing & Standing\_Clap \\
Standing\_Greeting & Standing\_Torch\_Idle\_02 \\
Standing\_Turn\_90\_Right & Standing\_Turn\_Left\_90 \\
Standing\_W\_Briefcase\_Idle & Stop\_Jumping\_Jacks \\
Strut\_Walking & Talking\_On\_Phone \\
Talking\_Phone\_Pacing & Talking  \\
Talking\_Turn\_180 & Walking  \\
Yawn & Yelling \\ 
\hline
\end{tabular}
\end{center}
}
\clearpage

\end{document}